\documentclass[final]{cvpr}
\usepackage{times}
\usepackage{epsfig}
\usepackage{graphicx}
\usepackage{amsmath}
\usepackage{amssymb}
\usepackage{color}
\usepackage{caption}
\usepackage{booktabs}
\usepackage[numbers,sort]{natbib}
\definecolor{hollywoodcerise}{rgb}{0.96, 0.0, 0.63}
\definecolor{lasallegreen}{rgb}{0.03, 0.47, 0.19}
\definecolor{hanpurple}{rgb}{0.32, 0.09, 0.98}
\definecolor{green(pigment)}{rgb}{0.0, 0.65, 0.31}
\usepackage[pagebackref=true,breaklinks=true,letterpaper=true,colorlinks,bookmarks=false]{hyperref}

\hypersetup{colorlinks,linkcolor={red},citecolor={hollywoodcerise},urlcolor={black}}

\def\cvprPaperID{3757} % *** Enter the CVPR Paper ID here
\def\confYear{CVPR 2021}
\begin{document}

%%%%%%%%% TITLE
% \title{\LaTeX\ Author Guidelines for CVPR Proceedings}
\title{EvDistill: Asynchronous Events to End-task Learning via Bidirectional Reconstruction-guided Cross-modal Knowledge Distillation}
% \and

\author{Lin Wang$^{1}$, Yujeong Chae$^{1}$\thanks{These two authors contributed equally.}, Sung-Hoon Yoon$^{1*}$, Tae-Kyun Kim$^{2}$,
and Kuk-Jin Yoon$^{1}$\\
$^{1}$Visual Intelligence Lab., KAIST, Korea\\
$^{2}$ICVL Lab., KAIST, Korea and Imperial College London, UK \\
{\tt\small \{wanglin, yujeong, yoon307, kimtaekyun,
kjyoon\}@kaist.ac.kr}
}

\maketitle

%%%%%%%%% ABSTRACT
\begin{abstract}
\vspace{-6pt}
%   Event cameras %\eg, DAVIS240, 
%   are 
% %   bio-inspired 
%   novel sensors that offer several advantages such as high frame rates and dynamic ranges over conventional cameras. 
%   in their high dynamic range, no motion blur, etc. 
Event cameras 
% , \eg, DAVIS240C, 
sense per-pixel intensity changes 
% at each pixel 
and produce asynchronous event streams with
high dynamic range and less motion blur, showing advantages over the conventional cameras. 
% It has been demonstrated that events are potential for learning many end-tasks, \eg, object recognition.
% They have a lot of advantages over conventional cameras, 
% However, 
A hurdle of training event-based models is the lack of large qualitative
% high-quality 
labeled data.
% However,  one challenge is the lack of labeled training data, which is key to training most recognition models because of their novelty. 
% However, 
%    
%   instead of fixed frame images
    % existing vision algorithms are less applicable and 
    % One obstacle is the lack of large labeled training data, 
    % % because of their novelty
    % which is key to training the end-task network models.
%  Prior-arts have reconstructed intensity images from asynchronous event streams in limited qualities.
%   What we are interested is to use events directly to end-tasks such as semantic image segmentation or object recognition. Note, however, there does not exist quality annotated event data or event-intensity image pairs. 
   %However, as the outputs are asynchronous `event' streams, 
%   instead of fixed frame images
    % existing vision algorithms are less applicable and 
    %qualitative annotated data for training deep neural network (DNN) models become challenging for achieving many end tasks. 
%   Although deep learning has been adapted to events and many approaches proposed to leverage simulated events; however, the domain gap with real events painfully hinders their potential and often leads to less plausible performance on many tasks.
%   Some attempts have been taken to leverage simulated event data; however, the domain gap with real-world events painfully hinders its potential and utility for handling many vision tasks. 
% %%%%motivation 
Prior works learning end-tasks mostly rely on labeled or pseudo-labeled datasets obtained from the active pixel sensor (APS) frames; however, such datasets' quality is far from rivaling 
% compared to 
those based on the canonical images.
% As some even cameras provide active pixel sensor (APS) frames, prior arts explored the pseudo labels based on APS frames to train the event data. However, as APS frames are of low resolution and quality, the pseudo labels are also not accurate.
In this paper,
% we address this limitation and 
 we propose a novel 
% image reconstruction-enhanced cross-modal knowledge distillation (CMKD)
approach, called \textbf{EvDistill}, to learn a student network on the unlabeled and unpaired event data (target modality) via knowledge distillation (KD) from a teacher network trained with large-scale, labeled image data (source modality).
% To enable KD across the unpaired modalities,
%  The teacher network is trained with abundant labeled image data (the source modality), and the student network on unlabeled and unpaired event data (the target modality). 
%to learn a student network on event modality by distilling the knowledge from a robust teacher network trained with abundant labeled image data.
 % We explore the potential of the real-world labeled image data and aim to learn a robust DNN model for the unlabeled event data inspired by cross-modal learning
%  transfer these knowledge of the learned models to the model 
%  learning events for end tasks.
% %  
%%%%%%probelms of existing cross-modal learning methods
%However, existing methods mostly rely on paired data (\eg, RGB and depth) with the same labels, which are ill-considered for our problem as there are no paired data of both modalities and only the labels for image modality are available.
%%%%%%%%%%%%proposed method
% to the network learning the unlabeled and unpaired target events on the same end task.
%   we propose a novel cross-modal knowledge distillation approach by exploring the potential of conventional camera images as there exist a large amount of labeled data and further transferring the knowledge from the learned models to the model learning event streams.
% To overcome this problem,
% As the source and target modalities are not paired,
To enable KD across the unpaired modalities, we first propose a bidirectional modality reconstruction (BMR) module to bridge
both modalities 
and simultaneously exploit them to distill knowledge via the crafted pairs, causing no extra computation in the inference. The BMR is improved by the end-tasks and KD losses in an end-to-end manner.
% , learnt in unsupervised manner.
% It is as part of the proposed entire framework, improved by end-tasks in end-to-end training. 
%First, although no paired data with labels for both modalities are available, we observe that one modality can be an intermediate visual representation of the other. We thus propose an bidirectional reconstruction scheme to connect the source and target modalities. 
% Using the reconstructed pairs of events and images, we perform knowledge distillation from the teacher to student network. 
Second, we leverage the structural similarities of both modalities and adapt the knowledge by matching their distributions. Moreover, as most prior feature KD methods are uni-modality and less applicable to our problem, we propose to leverage an affinity graph KD loss to boost the distillation.
% of EvDistill.
% APS (Active Pixel Sensor) frames obtained by an event camera is also incorporated to bridge the gap and distill better among the modalities. 
% The bidirectional reconstruction and knowledge distillation for end-tasks are learnt in end-to-end. 
Our extensive experiments on semantic segmentation and object recognition demonstrate that EvDistill achieves significantly better results than the prior works and KD with only events and APS frames.
% without using the BMR and source data.
% in the experiments of . 
% In the experiments of semantic segmentation and object recognition, our approach achieves significantly better results than the prior-arts and 
% % the basic setting of
% % for learning the end tasks and 
% % by pseudo labels 
% KD with only events and active pixel sensor (APS) frames without using the BMR and source data. 
% or the reconstructed intensity images. 
% and existing networks. 
%[DNN]

% both semantic segmentation and object recognition. 
%   As some event cameras, \eg, DAVIS 240, also offer APS images, we fully leverage these images as a medium of connecting conventional camera images via domain adaptation. To this end, we propose a cross-modal knowledge distillation framework that transfer the knowledge from conventional camera images (a.k.a, RGB modality) to event streams (event modality) learning the same visual task. 
%   To enhance the learning efficiency, we also propose a data augmentation module by reconstructing intensity images from events.
%   We conduct experiments on several high-level vision tasks: semantic segmentation, object recognition, and demonstrate the efficacy of our method.
   
\end{abstract}

\begin{figure}[t!]
    \centering
    \captionsetup{font=small}
    \includegraphics[width=.92\columnwidth]{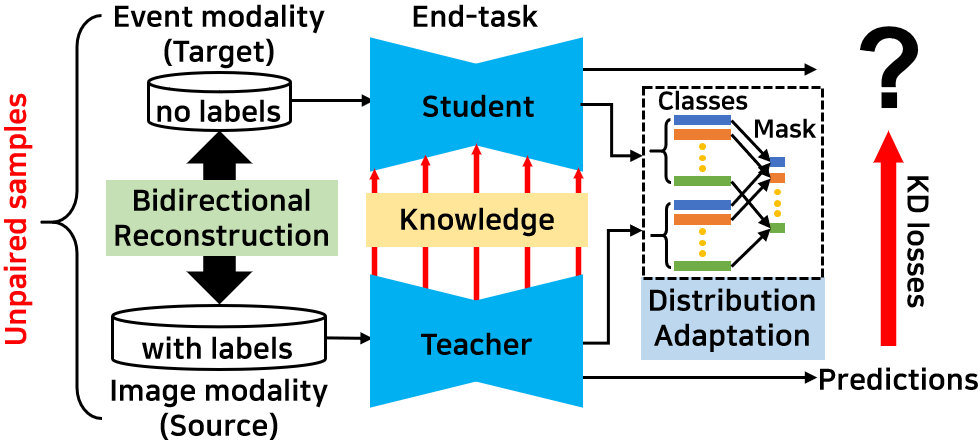}
    \vspace{-5pt}
    \caption{EvDistill distills knowledge from a teacher network trained with large labeled images to a student network learning unpaired and unlabeled events for the end-tasks. To distill knowledge, a bidirectional modality reconstruction and distribution adaptation schemes, with the novel KD losses, are proposed.}
    \label{fig:cover}
    \vspace{-12pt}
\end{figure}
%%%%%%%%% BODY TEXT
\vspace{-13pt}
\section{Introduction}
\vspace{-3pt}
% \noindent \textbf{Background and Motivation.}
% Event cameras, \eg, DAVIS240C \cite{brandli2014240}, are bio-inspired sensors that 
% are bio-inspired sensors that work radically different from the conventional cameras. 
% Compared to conventional cameras
% event cameras 
Event cameras %, \eg, DAVIS240C\cite{brandli2014240}, 
% also known as neuromorphic cameras, 
have recently received much attention in the computer vision and robotics community for their distinctive advantages, such as high dynamic range (HDR) and much less motion blur. Event cameras 
sense the intensity changes at each pixel asynchronously 
% at the time 
%when they occur 
% instead of a fixed frame rate. These changes 
and produce event streams encoding time, pixel location, and polarity (sign) of intensity changes.
Recently, deep neural network (DNN)-based methods with large-scale, labeled image data have shown significant performance gains on many tasks. % on the image-based cameras. 
However, learning effective event-based DNNs has been impeded by the lack of large 
% amounts 
pixel-level labeled event data. Prior works learning event-based high-level tasks have resorted to 
the manually annotated task-specific datasets in a supervised manner \cite{rebecq2019high, gehrig2020video, gehrig2019end,alonso2019ev, maqueda2018event, moeys2016steering, bi2019graph,cannici2019asynchronous,wang2019ev, tulyakov2019learning}. Although some labeled event datasets \cite{zhu2018multivehicle, orchard2015converting, hu2016dvs} 
% for learning end-tasks 
have been collected, the quantity and quality are far less favorable  compared to those based on the canonical images. Some works \cite{alonso2019ev, gehrig2020video} have made 
% have learned the end-task using 
pseudo labels using the active pixel sensor (APS) images; however, these labels are less accurate due to the low quality of APS images and considerable domain gap with the source data. While \cite{zhu2019unsupervised, zhu2018multivehicle} have explored unsupervised learning, they only focus on the pixel-level prediction tasks, \eg, depth estimation. Another line of research has reconstructed intensity images from events \cite{rebecq2019high,wang2019event,mostafavi2020learning, wang2020eventsr, wang2020event, stoffregen2020reducing}, and these images have been used to learn DNNs on end-tasks, \eg, object recognition \cite{rebecq2019high}; however, annotated labels are still needed, and extra latency is introduced in the inference time.                                                
We explore to leverage large labeled image data (a.k.a. source modality)
% (as there are abundant annotations) 
and the learned models, and aim to learn a model on the unpaired and unlabeled events (a.k.a. target modality) via cross-modal learning \cite{gupta2016cross, zhao2020knowledge}  and knowledge distillation (KD) \cite{wang2020knowledge, gupta2016cross,thoker2019cross}. 
Most existing cross-modal learning methods have relied on paired data (\eg, image and depth) with the same labels \cite{gupta2016cross,wang2019efficient, zhao2020knowledge,yuan2019ckd,hafner2018cross,li2020towards,xu2018pad,owens2016ambient} or extra information (\eg, data or labels)  \cite{garcia2018modality,yuan2018rgb,baek2020weakly, rad2018domain} or grafting networks  between modalities (\eg, image to thermal) \cite{hu2020learning} for learning the end-tasks. 
% Although \cite{hu2020learning} have grafted networks from image modality to another unlabeled modality (\eg, thermal), paired data are required.
Some works have explored the unpaired multi-modality data \cite{dou2020unpaired, li2020towards}; however, it is assumed that labels for both modalities are available, which is difficult to achieve for the event data.  
% While several methods exploited the privileged information \cite{garcia2018modality,yuan2018rgb} or domain adaptation \cite{baek2020weakly, rad2018domain} for the paired modalities, side information or additional labels are required.
% for weak supervision. 
% Moreover, some approaches explored adversarial learning
% % for cross-modal information
% \cite{hu2020creating, dou2020unpaired, li2020towards} for the unpaired multi-modal learning; however, these methods assume that end-task labels for both modalities are available.
% First of all, we observe that although events and images are two different modalities, they are some how inter-connected to the same scene perception. That is, even when labeled event data are difficult to acquire, this challenge could be possibly relieved by fully leveraging the conventional images as there exist a large amount of labeled data. Secondly, there exist a large amount of the state-of-the-art (SoTA) network models trained on various annotated image datasets. Therefore, it could be possible to transfer the knowledge of a learned network on labeled images to the network learning the unlabeled events.
% In this paper, we address the shortage of event-camera data, especially the annotated data for training, and further exploit potential and inter-connections of both events and APS frames. 
% Inspired by the methods for cross-modal learning \cite{wang2020knowledge, gupta2016cross, zhao2020knowledge, thoker2019cross, yuan2018rgb, rad2018domain,baek2020weakly}, 
%

To overcome these limitations, we propose a novel method, called \textbf{EvDistill}, to efficiently learn a student network on the unpaired and unlabeled event data by distilling the knowledge from a robust teacher network trained with large labeled image data, as shown in Fig.~\ref{fig:cover}. Firstly, we propose a bidirectional modality reconstruction (BMR) module to bridge both modalities, and then simultaneously exploit them to distill knowledge via the crafted pairs, adding no extra computation cost during inference (Sec.~\ref{data_augmentation}). Importantly, the BMR is improved by the end-task and the KD losses in an end-to-end manner. That is, BMR produces the crafted pairs of both modalities 
to distill knowledge to the student network in the \textit{forward} pass, and KD facilitates the learning of the BMR in the \textit{backward} pass.  
% although no paired data with labels for both modalities are available, 
% we observe that one modality could be an intermediate representation for the other, 
% In such a way, paired data can be crafted and labels in source modality can be fully utilized. 
Secondly, as the feature representations of two modalities extracted from the task networks could suffer from distribution mismatch,
%  As some event cameras,\eg, DAVIS240 \cite{brandli2014240}, offer active pixel sensor(APS) images and some data of both modalities contain similar spatial structures, \eg, road driving,
 we leverage the structural similarities and adapt knowledge by matching the class distributions based on the BMR module (Sec.~\ref{domain_adap}). Moreover, as most existing methods are limited to uni-modality \cite{romero2014fitnets,kim2018paraphrasing, zagoruyko2016paying,liu2019structured}, we leverage a graph affinity KD loss and other losses to learn a better model on the event data (Sec.~\ref{loss_terms}).
We evaluate the performance of the proposed framework on three datasets in semantic segmentation (Sec.~\ref{ev_seg_sec}) and one dataset in object recognition (Sec.~\ref{cls_kd}). The experiments show that our approach achieves significantly better performance 
% ($9.5\%$ 
% higher on semantic segmentation 
% and $6\%$, respectively) 
than the prior works for both end-tasks and the naive setting, KD with only events and the APS frames (when APS frames are available). 
% without using the BMR module and source data
The validation code and trained models are available at \url{https://github.com/addisonwang2013/evdistill}.
\vspace{-5pt}
\section{Related Works}
\vspace{-4pt}
% mainly talk about event to image recon. and  segmentation, recognition, simulated events
\noindent \textbf{DNNs for event-based end-tasks.} DNNs with event data was first explored for the classification task \cite{neil2016phased} and for robot control \cite{moeys2016steering}. \cite{maqueda2018event} then trained a DNN for steering angle prediction on DDD17 dataset \cite{binas2017ddd17}. This dataset has been utilized by \cite{alonso2019ev,gehrig2020video} to perform semantic segmentation using pseudo labels obtained from the APS frames. Moreover, DNNs have been applied to some high-level prediction tasks, such as %optical flow estimation \cite{zhu2019unsupervised, gehrig2019end, stoffregen2020train,gallego2019focus}, stereo depth estimation \cite{zhu2019unsupervised, tulyakov2019learning} on MVSEC dataset \cite{zhu2018multivehicle}, 
object detection and tracking \cite{cannici2019asynchronous,hu2020learning,messikommer2020event,ramesh2018long}, human pose estimation \cite{wang2019ev,xu2020eventcap,calabrese2019dhp19}, motion estimation \cite{stoffregen2019event,mitrokhin2020learning,kepple2020jointly,wang2020stereo,yang2020vess}, 
object recognition \cite{gehrig2020video, rebecq2019high, bi2019graph} on N-Caltech \cite{orchard2015converting} and other datasets \cite{sironi2018hats,li2017cifar10,bi2019graph}.  \\
% \vspace{-3pt}
\noindent \textbf{DNNs for event-based  low-level vision.} 
% Meanwhile, another line of research focuses on the low-level vision tasks. 
% DNNs have been applied to 
Meanwhile, another line of research focuses on the low-level prediction tasks, such as optical flow estimation \cite{zhu2019unsupervised, gehrig2019end, stoffregen2020reducing,gallego2019focus}, stereo depth estimation \cite{zhu2019unsupervised, tulyakov2019learning} on MVSEC dataset \cite{zhu2018multivehicle}. In addition, 
\cite{wang2019event, rebecq2019high, paredes2020back,scheerlinck2020fast,stoffregen2020reducing,mostafavilearning} attempted to reconstruct intensity image/video from events using camera simulator \cite{rebecq2018esim, mueggler2017event}, and \cite{wang2020eventsr, mostafavi2020learning,wang2020event} tried to reconstruct high-resolution images. In contrast to image/video reconstruction from events, \cite{gehrig2020video} proposed to generate events from video frames. Some other works also explored the potential of events for image deblurring \cite{haoyu2020learning,jiang2020learning, wang2020eventsr}, HDR imaging \cite{han2020neuromorphic,zhang2020learning}, and event denoising \cite{baldwin2020event}.
% Moreover, \cite{baldwin2020event} exploited deep learning for event denoising.
% trained CNNs for robot control using both event streams and APS images, 
% \cite{alonso2019ev} further used an encoder-decoder structure for event-based segmentation on DDD17 dataset \cite{binas2017ddd17}. 
% In contrast, \cite{zhu2019unsupervised} utilized an encoder-decoder network for optical flow, depth and ego-motion estimation via unsupervised learning. 
% Besides, \cite{cannici2019asynchronous} refined YOLO \cite{redmon2016you} for event-based object detection. \cite{tulyakov2019learning} represented events as a composition of temporal and spatial aggregations and showed good performance on stereo matching. 
% Meanwhile,\cite{gallego2019focus} proposed some loss functions to align event focus and \cite{gehrig2019end} designed an end-to-end event representation framework, both of which are further applied to flow estimation, etc. 
%  Moreover, \cite{baldwin2020event, jiang2020learning} focused on event-based denoising and debluring, respectively. 
 For more details about event-based vision, refer to a survey \cite{Gallego2020EventbasedVA}. Differently, we propose EvDistill, learning event-based end-tasks on the unpaired and unlabeled events via cross modal KD,
 in which a BMR moudle is proposed to bridge both modalities and is learned with the end-task networks in an end-to-end manner.
%  which is achieved by the bidirectional modality reconstruction and distribution adaptation schemes with novel KD losses.  
% They also used an event sensor with VGA ($640 \times 480$ pixels) resolution to reconstruct higher resolution video, however, the problem is essentially different from our work. 
% as proposed in \cite{Gallego_2018_CVPR}
% . They further apply these loss functions to rotational motion, depth and optical flow estimation \cite{zhu2019unsupervised}. 
% Similar optimization approaches are also proposed by Stoffregen \etal \cite{stoffregen2019event}. 
% They propose some reward functions to handle aperture problems, noise and data insufficiency, and validate their method via optical flow estimation.

\noindent \textbf{Knowledge Distillation.} KD  aims to build a smaller (student) model with the softmax labels of a larger (teacher) model \cite{romero2014fitnets,hinton2015distilling,wang2020knowledge}. 
% KD is usually characterized by its so-called `Student-Teacher' learning . 
Most KD methods learning end-tasks have been focused on uni-modality (\eg, image) data and distill knowledge using logits \cite{chen2019online,zhang2018deep,xu2020knowledge,yao2020knowledge} or features \cite{romero2014fitnets, kim2018paraphrasing, zagoruyko2016paying,park2019feed,heo2019comprehensive}. Cross-modal KD aims to transfer knowledge across different modalities. Most prior cross-modal KD methods \cite{gupta2016cross,wang2019efficient, hafner2018cross,zhao2020knowledge,hu2020creating,yuan2019ckd, chen2019synergistic,li2020towards,xu2018pad, dou2020unpaired,pande2019adversarial,owens2016ambient,hu2020learning} relied on the paired data with the common labels, while some works utilized extra information (\eg, data) \cite{garcia2018modality,yuan2018rgb,baek2020weakly, rad2018domain} or grafting networks \cite{hu2020learning} to transfer knowledge. Although \cite{dou2020unpaired, li2020towards} explored the unpaired multi-modal KD, the labels for both modalities are needed.
% However, the capacity of knowledge in this setting is quite limited. To overcome this issue,
% Meanwhile, some works, \eg, \cite{chen2019online}, focused on the ensemble of logits information of multiple teachers to help the student in contrast to \cite{park2019feed} focused on the ensembled features based on the loss in \cite{zagoruyko2016paying, kim2018paraphrasing}. Some methods explored mutual \cite{zhang2018deep}
%  or adversarial learning \cite{belagiannis2018adversarial}. 
%  Recently, some works \cite{gupta2016cross,wang2019efficient, hafner2018cross,zhao2020knowledge,hu2020creating,yuan2019ckd,hu2020creating, chen2019synergistic,hafner2018cross,li2020towards,xu2018pad, dou2020unpaired,pande2019adversarial} extended KD by applying it to transfer knowledge across different modalities (RGB image and depth) using the paired data with common labels on the end-tasks. 
% Moreover, some approaches explored adversarial learning
% % for cross-modal information
% \cite{hu2020creating, dou2020unpaired, li2020towards} for the unpaired multi-modal learning; however, these methods assume that end-task labels for both modalities are available.
For more details about KD, refer to %a survey paper 
\cite{wang2020knowledge}. 
%  Differently,
%  Different from these works utilizing paired data of two modalities with the same labels, 
Learning from event data is more challenging as no paired labels for two modalities exist; we thus propose EvDistill, where we
%  , and only the labels of image are available; we thus 
%  As one modality could be an intermediate representation of the other, 
exploit a BMR module to connect them and simultaneously distill knowledge by adapting distribution with novel KD losses. 

\begin{figure*}[t!]

    \centering
     \captionsetup{font=small}
    \includegraphics[width=0.96\textwidth]{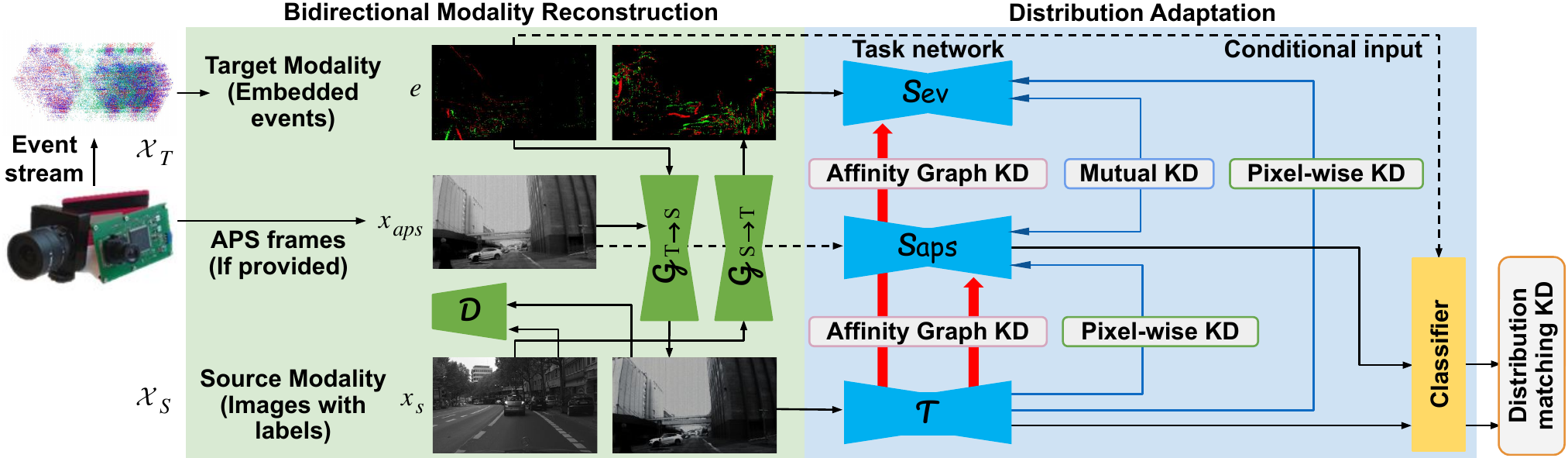}
    \vspace{-8pt}
    \caption{{Overview of the proposed EvDistill framework}. The architecture comprises a teacher network $\mathcal{T}$ and two student networks $\mathcal{S}_{ev}$ and $\mathcal{S}_{aps}$ (when APS frames exist). As there is no paired data with the same labels for both modality, a novel bidirectional reconstruction module is proposed to connect image and event modalities. Meanwhile, a distribution adaptation scheme with novel KD losses is also proposed to match the spatial structural distribution of both modalities.}
    \label{fig:kd_method}
    \vspace{-11pt}
\end{figure*}

\vspace{-3pt}
\section{The Proposed Method: EvDistill}
\vspace{-3pt}
\noindent \textbf{Event Representation}
\vspace{-3pt}
\label{event_rep}
 An event $e$ is interpreted as a tuple $(\textbf{u}, t, p)$, where $\textbf{u}= (x,y)$ is the pixel coordinate, $t$ is the timestamp, and $p$ is the polarity indicating the sign of brightness change. An event occurs whenever a change in log-scale intensity exceeds a threshold. 
%  To process event streams using DNNs, it is required to stack sparse events into image-like or a fixed tensor-like representations \cite{rebecq2019high, wang2019event, gehrig2019end}. 
% An event camera interprets the intensity changes as asynchronous event streams. 
% \begin{equation}
%     L(x,y,t) - L(x,y, t-\Delta t) \ge pC
% \end{equation}
% where $p \in \{-1,1\}$, and $\Delta t$ is the time interval since the last event at pixel $\textbf{u}={x,y}$. 
% A number of events are triggered in a given time interval $\Delta t$, which can be denoted as:
% \begin{equation}
%     \mathcal{E} = e_{k=1}^{N} =\{\textbf{u}_k, t_k, p_k \}_{i=1}^{N}
% \end{equation}
A natural choice is to encode events in a spatial-temporal 3D volume to a voxel grid \cite{rebecq2019high,zhu2019unsupervised, zhu2018unsupervised} or event frame \cite{rebecq2017real, gehrig2019end} or multi-channel images \cite{wang2020event, lin2020learning, wang2019event}. In this paper, we represent events to multi-channel event images as the inputs to the DNNs. Details are provided in the suppl. material.
% For semantic segmentation, we also consider the 6-channel representation \cite{alonso2019ev}.

% The duration of an event stream is $\Delta T = t_{N-1} -t_0$. For convenience, the events are discretized into $B$ temporal bins. The voxel is described as:
% \vspace{-6pt}
% \begin{equation}
%     \mathcal{E}(x,y, t)= \sum p_i \max(0, 1-|t-t_i^*|)
% \end{equation}
% where $t_i^* =\frac{B-1}{\Delta T} (t_i-t_0)$ is the normalized timestamp. To better facilitate learning, we represent the grid into single- or multi-channel event images. For semantic segmentation, we also consider the 6-channel representation \cite{alonso2019ev}.
% representation of events  
% For first one, we accumulate two bins of events and divide the positive and negative events to two channels. For the second one, we accumulate five bins of events and form a event frame.  
% \subsection{Cross-Modal Knowledge Distillation for Events}

\vspace{-1pt}
\subsection{Overview}
\vspace{-3pt}
% \noindent \textbf{Overview.} 
We describe the proposed EvDistill framework for learning end-tasks from events, as shown in Fig.~\ref{fig:kd_method}. 
For event cameras, \eg, DAVIS346 \cite{taverni2018front} with APS frames, assume that we are given the target modality data $\mathcal{X}_T = \{e, x_{aps}\}_i$ \textit{without labels}, where $e_i$ and $x_{{aps}_i}$ are $i$-th embedded event image and corresponding APS image. However, the source modality image data $\mathcal{X}_S= \{x_{s}, y_{s}\}_j$ are \textit{with labels}, where $x_{s_j}$ is $j$-th image with its label $y_{s_j}$. Unlike prior cross-modal learning methods \cite{zhao2020knowledge, gupta2016cross}, we assume there are no paired source and target modality data with common labels. 
% Moreover, unlike other modality data, \eg, depth, event data are very sparse and noisy, which makes it difficult to directly apply existing cross-modal KD methods.  
We address the challenge by proposing EvDistill, where $\mathcal{X}_S$ and $\mathcal{X}_T$ are not paired, and only the labels of $\mathcal{X}_S$ are available. In EvDistill, there are two student networks $\mathcal{S}_{ev}$ for learning events and $\mathcal{S}_{aps}$ for learning APS images (when APS images are provided). 
% During training, $\mathcal{S}_{aps}$ and $\mathcal{S}_{ev}$ not only learn from the same teacher $\mathcal{T}$  but also learn from each other.
Our goal is to train a student network $\mathcal{S}_{ev}$ learning events by distilling knowledge from a teacher network $\mathcal{T}$.
% on labeled image modality data. 
% As mentioned above, the main challenge is that $D_S$ and $D_T$ are not paired and only the labels of $D_S$ are available. 
Our key ideas are three folds. First, as data of both modalities $\mathcal{X}_S$ and $\mathcal{X}_T$ are unpaired,
% , there might exist a way of connecting both modalities . We 
we thus propose a bidirectional reconstruction module to bridge both modalities and then simultaneously exploit them to distill knowledge via the crafted pairs (Sec.~\ref{data_augmentation}). Second, as there exist spatial structure similarities (\eg, cars, people in urban scenes) between the two modalities, we propose a distribution adaptation scheme to adapt the knowledge by matching the class distribution of the two modalities (Sec.~\ref{domain_adap}). Lastly, as some end-tasks, \eg, semantic segmentation, aim to predict pixel-wise class information, we propose a novel affinity graph KD loss and employ other loss terms to learn a better $\mathcal{S}_{ev}$ (Sec.~\ref{loss_terms}). 
% We now describe the technical details in the following sections.

% \vspace{-5pt}
% to generalize the learned knowledge from $D_S$ to $D_T$. This is achieved by two approaches. 
% First, the event modality can be interpreted as image modality via semantics-enhanced image reconstruction . Second, when APS frames are available, the knowledge can be generalized by minimizing the class distribution gap with the source images (Sec.~\ref{domain_adap}). Third, we propose a novel graph-based KD bridge together with mutual learning \cite{zhang2018deep} and adversarial learning to improve the distillation of knowledge to events (Sec.~\ref{ckmd_losses}). In the following sections, we will describe these approaches in details.
% there is a source modality (image) and a target modality (event stream). Our method teacher performs domain adaptation (DA) of a teacher model (RGBNet) by learning domain invariant features between the RGB images with labels and the APS images from an event camera. Meanwhile, it progressively distills the knowledge its knowledge to a student model one (IntensityNet) learning APS images and a student model two learning event streams (EventNet) on both source and target features.  As shown in Fig.~\ref{fig:kd_method}, DA is achieved on the teacher network and the student one, while the KD from teacher to the student one and student two. To further improve both KD and DA, we also propose to add an data augmentation module. 
% \vspace{-10pt}
% \subsection{Bidirectional Modality Reconstruction}
\vspace{-5pt}
\subsection{Bridging Source and Target Modalities}
\vspace{-3pt}
\label{ckmd_losses}
% Learning event data without labels is a challenging task. To  enable efficient learning of event data, meanwhile, relieve the burden of obtaining sufficient ground truth labels, we propose a novel cross-modal knowledge distillation framework, as shown in Fig.~\ref{fig:kd_method}. 
% In our KD framework,  there are two student networks $\mathcal{S}_{ev}$ for learning events and $\mathcal{S}_{aps}$ for learning APS frames (when APS frames are provided). During training, $\mathcal{S}_{aps}$ and $\mathcal{S}_{ev}$ not only learn from the same teacher $\mathcal{T}$  but also learn from each other.
% On one aspect, as the teacher network $\mathcal{T}$ contains significant amount of knowledge, we aim to distill these knowledge to both the $\mathcal{S}_{ev}$ and $\mathcal{S}_{aps}$. However, as there are no paired data from both modalities $D_S$ and $D_T$ and only the labels of $D_S$ are available, it is imperative that we need to explore a way to generalize the knowledge in such a challenging situation. We thus propose two approaches to tackle the challenge in the following sections.
% The intuition behind this is that there exist large amount of labeled dataset and a large number of state-of-the-art(SoTA) network models for image modality. 
% Empowered by the unsupervised domain adaptation module, as depicted in Sec.~\ref{domain_adap}, the student $\mathcal{S}_1$ and student $\mathcal{S}_2$ can learn sufficient knowledge from $\mathcal{T}$. 

% \noindent \textbf{Semantics-enhanced Event to Image Reconstruction }
\label{data_augmentation} 
Although data from both modalities are unpaired, we observe that one modality could be the alternative representation of the other modality under the same end-task. 
% Recently, some attempts have been taken on image reconstruction \cite{rebecq2019high, wang2019event, wang2020eventsr, mostafavi2020learning, scheerlinck2020fast}; however, these methods are mostly supervised and are optimized by the pixel-wise loss. For EvDistill, there are no paired data from both modalities. 
We thus propose a bidirectional modality reconstruction (BMR) module to bridge both modalities for enabling distillation on the $\mathcal{S}_{ev}$. As shown in Fig.~\ref{fig:kd_method}, the proposed BMR consists of two generators where generator $\mathcal{G}_{T \to S}$ translates events to an intermediate representation in the image modality, and $\mathcal{G}_{S \to T}$ translates the labeled image data to an intermediate representation in the event modality.  In such a way, the labels of image modality can be leveraged as supervision on the intermediate representation in the event modality when training $\mathcal{S}_{ev}$. Meanwhile, the intermediate image representation of events also helps to learn the knowledge (\eg, predicted labels) from the teacher $\mathcal{T}$.
% As a potential, it could help bridge both modalities and generalize the knowledge for learning $\mathcal{S}_{ev}$. The reconstructed images are expected to contain the same semantic information as the intensity images. 
% However, as $x_{aps}$ and source image $x_s$ are from different image domains, and it might be 
% Thus, the reconstructed images can be leveraged as the `cohort' of events for achieving the learning tasks.
% As there exist no paired source and target modality data and labels
% Inspired by some recent attempts \cite{rebecq2019high, wang2019event, wang2020eventsr,mostafavi2020learning}, we propose a semantics-enhanced event to image reconstruction module $\mathcal{T}$ network to the student network $\mathcal{S}_{ev}$ and the student network $\mathcal{S}_{aps}$. 
% However, directly applying existing frameworks may lead to less optimal performance as generated intensity images are expected to contain the same semantic information as original intensity images for facilitating dense prediction tasks, \eg, semantic segmentation. Following this clue, we introduce a semantic-enhanced event to image reconstruction module (SEIR). The SEIR not only bridges events and APS frames but also connect the $\mathcal{T}$ network in an end-to-end learning manner.
% In our framework, we propose a data augmentation module based on GAN \cite{goodfellow2014generative, wang2019event}.
% For simplicity, we employ the student network $\mathcal{S}_{aps}$ trained on
% the adapted domain as $\mathcal{T}$, i.e. $\mathcal{S}_{aps}$ = $\mathcal{T}$, 
% \noindent \textbf{Forward.}
When the APS frames are available in some event cameras, %\eg, DAVIS240C, 
we utilize APS frames and apply 
% a perceptual loss based on the LPIPS (Learned Perceptual Image Patch Similarity) \cite{zhang2018unreasonable} instead of
the pixel-wise loss for the supervision of $\mathcal{G}_{T \to S}(e)$. The generated images are further adapted to the source data $\mathcal{X}_S$. The pixel-wise loss is defined as:
% can be formulated as:
\vspace{-2pt}
\begin{equation}
    \mathcal{L}_{BMR}^{pw} = \mathbb{E}_{e, x_{aps} \sim \mathcal{X}_T}[||x_{{aps}} - \mathcal{G}_{T \to S}(e)||_1]. \\[-2pt]
\end{equation}
% where $\Phi(\dot)$  is the intermediate layer of a pretrained network (\eg, VGG).
Moreover, BMR is enhanced by the cycle consistency loss \cite{zhu2017unpaired} and adversarial loss, which are crucial for the mapping of the two modalities. The adversarial loss (\eg, from target to source modality) is formulated based on \cite{goodfellow2014generative,wang2020deceiving}:
\vspace{-2pt}\begin{equation}
\mathcal{L}_{BMR}^{Adv} = \mathbb{E}_{e \sim \mathcal{X}_T}[1- \log(\mathcal{D}(\mathcal{G}_{T \to S}(e))],  \\[-2pt] 
\end{equation}
where $\mathcal{G}$ is the generator and $\mathcal{D}$ is the discriminator.

\begin{figure}[t!]
    \centering
     \captionsetup{font=small}
    \includegraphics[width=\columnwidth]{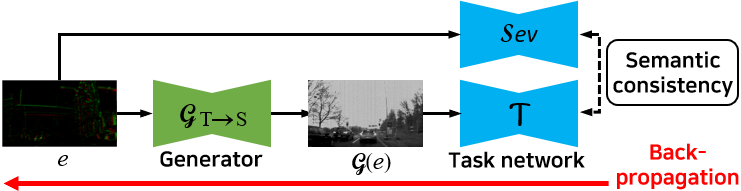}
    \vspace{-18pt}
    \caption{Illustration of the end-to-end target-to-source modality reconstruction $\mathcal{G}_{T \to S}$ with task net $\mathcal{T}$ and $\mathcal{S}_{ev}$ in the BMR module. }
    \label{fig:image_syns}
    \vspace{-15pt}
\end{figure}

\vspace{2pt}
\noindent \textbf{End-to-end learning.} 
To better preserve semantic information, we exploit a novel dynamic semantic consistency (DSC) loss and build our framework based on adversarial learning \cite{goodfellow2014generative, wang2019event,zhao2019multi}, as shown in Fig.~\ref{fig:image_syns}. The proposed DSC loss for BMR has three advantages: (1) the generated intermediate representation of events in image modality $\mathcal{G}_{T \to S}(e)$ becomes the optimal input of $\mathcal{T}$ and the generated intermediate representation of source images in event modality $\mathcal{G}_{S \to T}(x_s)$ becomes the optimal input of $\mathcal{S}_{ev}$; (2) the generated intermediate representations $\mathcal{G}_{T \to S}(e)$ and $\mathcal{G}_{S \to T}(x_s)$ both provide the supervision for $\mathcal{S}_{ev}$ based on the knowledge of $\mathcal{T}$ and labels of image data; (3) importantly, the BMR module is improved by these constraints on the end-tasks (\eg, cross-entropy loss) in an end-to-end manner (see Fig.~\ref{fig:image_syns}). That is, the KD loss promotes learning of the BMR module in the \textit{backward} pass, and the BMR module produces crafted pairs to distill knowledge to the student network $\mathcal{S}_{ev}$ in the \textit{forward} pass, which will be described in Sec.~\ref{domain_adap}. \textit{The BMR module
can be removed after training, leading to no additional computation cost during inference time}. The proposed DSC
% dynamic semantic consistency (DSC) 
loss for, \eg, target-to-source modality learning is as: 
% \vspace{-2pt}
\begin{equation}
\label{kl_div}
\begin{split}
   \mathcal{L}_{BMR}^{DSC} = \mathbb{E}_{e \sim \mathcal{X}_T} KL[\mathcal{T}(\mathcal{G}_{T \to S}(e))||\mathcal{S}_{ev}(e)], 
%   + \\ \mathbf{E}_{x_s \sim \mathcal{X}_S}KL[\mathcal{T}(x_s)||\mathcal{S}_{ev}(\mathcal{G}_{S \to T}(x_s))]   
\end{split}
\end{equation}
where $KL$($\cdot||\cdot$) is the KL divergence between two distributions.
More detailed formulation of the proposed BMR and its total loss $\mathcal{L}_{BMR}$ is provided in the suppl. material.
% In summary, the overall loss for SSIM is:
% \begin{equation}
%     \mathcal{L}_{SEIR} = \mathcal{L}_{GAN} + \lambda_1 \mathcal{L}_{SC}  + \lambda_2 \mathcal{L}_{1}
% \end{equation}

\subsection{Distillation via Distribution Adaptation (DA)} 
\label{domain_adap}
\begin{figure}[t!]
    \centering
     \captionsetup{font=small}
    \includegraphics[width=.98\columnwidth]{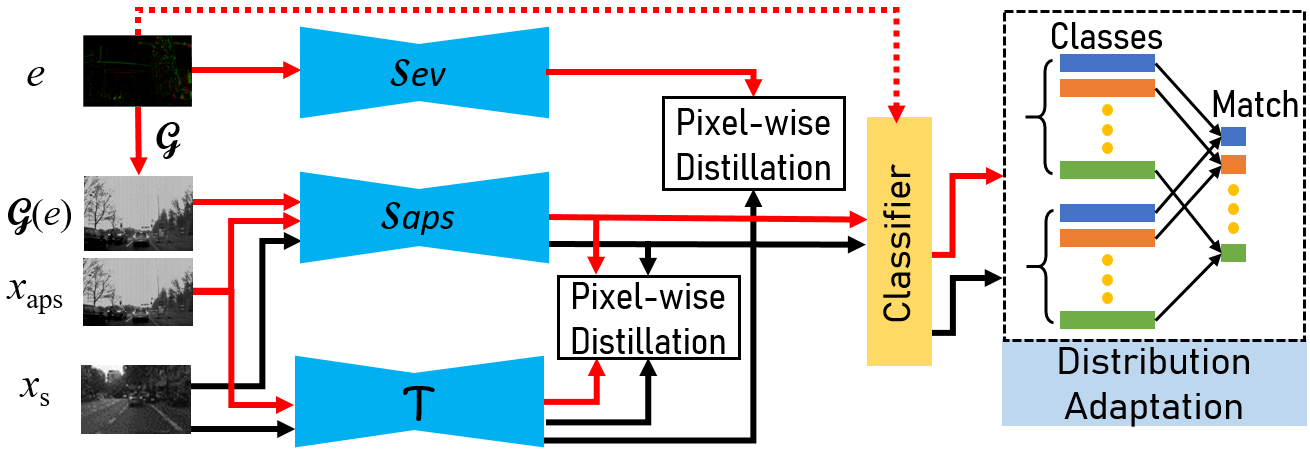}
    \vspace{-10pt}
    \caption{Illustration of distribution matching of the proposed distribution adaption scheme.}
    \label{fig:ukg_method}
    \vspace{-10pt}
\end{figure}

\vspace{-2pt}
% As some event cameras, such as DAVIS240, provides APS frames, it is of great importance to fully utilize APS images to better improve the learning on events. 
% However, we observe that there exists significant gap within image modality between $x_{aps}$ and $x_s$, 
% Even with the proposed BMR module, 
Based on the BMR module, the source and target modalities are connected, and we then further simultaneously exploit them to distill knowledge.
Due to the distinct difference between event
and image modality data, the features of two modalities extracted from the teacher and student networks suffer from the distribution mismatch.
% as demonstrated in the experiment. 
% This might degrade the distillation of knowledge from $\mathcal{T}$ to $\mathcal{S}_{ev}$. 
To address this issue, we propose to leverage the intrinsic spatial structure between source and target modality data $\mathcal{X}_S$ and $\mathcal{X}_T$. Our motivations are two folds. Firstly, as shown in Fig.~\ref{fig:ukg_method}, when APS frames are available, we propose to employ KD losses to guide the student $\mathcal{S}_{aps}$ to behave like the teacher $\mathcal{T}$ in addition to the cross-entropy (CE) loss $\mathcal{L}_{CE}$ based on source labels.
% it is worthwhile to align the pixel-wise predictions of the networks
% between $D_S$ and $D_T$, 
This is done by aligning both $\mathcal{G}_{T \to S}(e)$ and $x_{aps}$ with $x_s$ to generalize the feature information. 
% As the APS images from event cameras are shown to have domain gap with the source images  \cite{gehrig2020video, wang2020eventsr}, it is of great importance to first minimize the domain gap.
That is, the intermediate representation of events $\mathcal{G}_{T \to S}(e)$ and APS image $x_{aps}$, the source image $x_{s}$ are all fed to the teacher $\mathcal{T}$ and the student $\mathcal{S}_{aps}$ to match the features. For simplicity, we model prediction matching loss based on the pixel-wise loss (\eg, $l_1$), which is formulated as:
% , which can be formulated as:
\begin{equation}
\begin{split}
    \mathcal{L}^{aps}_{DA} = \mathbb{E}_{x_{aps} \sim \mathcal{X}_T}KL[\mathcal{S}_{aps}(x_{aps})||\mathcal{T}(x_{aps})]  + \\ \mathbb{E}_{x_{s} \sim \mathcal{X}_S}KL[ \mathcal{S}_{aps}(x_{s})||\mathcal{T}(x_{s})]
    % + \mathbf{E}||\mathcal{T}(\mathcal{G}_{T \to S}(e)) - \mathcal{S}_{aps}(\mathcal{G}_{T \to S}(e))||_1
    % \mathcal{L}^{UKG}_{D_S} = || \mathcal{T}(x_{s}) - \mathcal{S}_{aps}(x_{s})||_1
\end{split}
\end{equation}
We then propose to employ a distillation loss to guide the student $\mathcal{S}_{ev}$ to behave more like the teacher $\mathcal{T}$ based on the event $e$, which can be formulated as:  
\begin{equation}
\begin{split}
    %   \mathcal{L}_{KD}^{pw} = \alpha ||T(x_{aps}) - S_{aps}(x_{aps})||_1 +
    %   \\
    \mathcal{L}_{DA}^{ev}= \mathbb{E}_{e, x_{aps} \sim \mathcal{X}_T}[KL[\mathcal{T}(x_{aps})||\mathcal{S}_{ev}(e))] 
    % + 
    % \mathcal{L}_{BMR}^{DSC} 
    % +  \mathbf{E}KL[\mathcal{T}(\mathcal{G}_{T \to S}(e))||\mathcal{S}_{ev}(e)] + \mathbf{E}KL[\mathcal{T}(x_s) ||\mathcal{S}_{ev}(\mathcal{G}_{S \to T}(x_s))]
\end{split}
\end{equation}

However, the source image-guided distillation can not fully reduce the distribution mismatch as pixels may vary in either the appearance or scales from $\mathcal{X}_T$ and $\mathcal{X}_S$. For instance, cars are always small, and buildings are always large in either event or image modality. We then propose 
% a distribution adaption scheme by 
to directly match the distribution of class categories between the two modalities, as shown in Fig.~\ref{fig:ukg_method}. Specifically, denote $h : x \to \{0, 1\}$ as a
classifier, which is used to predict which modality an input pixel-level feature comes from, where 0 denotes the
source modality $\mathcal{X}_S$, and 1 denotes the target modality $\mathcal{X}_T$. Intuitively, training a distribution classifier is to distinguish
samples from two modalities. 
% To reduce the distribution difference, 
We encourage the activation $x$ to be modality indistinguishable. Considering each $x$ is generated from a task network, denoted by $\mathcal{F}$ (either $\mathcal{T}$ or $\mathcal{S}_{ev}$/$\mathcal{S}_{aps}$), we thus need to optimize $\mathcal{F}$
such that the distribution classification loss $\mathcal{L}_{DA}^{match}$ is maximized.
By jointly learning the distribution classifier $h$ and the task network $\mathcal{F}$, we arrive at the following maxmin problem, which can be optimized in an adversarial training manner \cite{goodfellow2014generative}. 
\vspace{-3pt}
\begin{equation}
    \mathcal{L}_{DA}^{match}(\mathcal{X}_S, \mathcal{X}_T)= \frac{1}{\mathcal{|X|}}\sum_{x \in \mathcal{X}}\mathcal{H}(h(x),d) \\[-3pt]
\end{equation}
Here, $\mathcal{X}= \mathcal{X}_S \cup \mathcal{X}_T$, $|\mathcal{X}|$ is the number of samples, $d \in \{0,1\}$ is the modality label, and $\mathcal{H}(\cdot)$ is a classification loss. For convenience, we adopt the conditional adversarial learning \cite{gulrajani2017improved, chen2018road} to optimize the matching problem. Finally, the distribution adaptation (DA) loss $\mathcal{L}_{DA}$ is the linear combination of the three loss terms  $\mathcal{L}_{DA}^{aps}$, $\mathcal{L}_{DA}^{ev}$ and $\mathcal{L}_{DA}^{match}$.
% The more details of data augmentation are given in Sec.\ref{data_augmentation}. With the DA module, the events are used to reconstruct intensity images. Denote $\mathcal{G}$ as the image reconstruction network and given stacked events as $e$. We can attain the reconstructed image $x_{
% gen} = \mathcal{G}(e)$. In such a way, additional DA losses are added to both $\mathcal{T}$ and $\mathcal{S}_{aps}$, which are formulated as:

% \begin{equation}
%   \mathcal{L}^{}_{Gen} = || \mathcal{T}(x_{gen}) - \mathcal{S}_{aps}(x_{gen})||_1 
% \end{equation}

% To facilitate to effectiveness of domain adaptation, we propose to add an data augmentation module, inspired by the recent attempts on intensity image reconstruction from events \cite{ rebecq2019high, wang2019event, wang2020eventsr, mostafavi2020learning}. 

% \noindent \textbf{Pixel-wise distillation.} To achieve the goal of cross-modal distillation from $\mathcal{T}$ to the student $\mathcal{S}_1$ and the student $\mathcal{S}_2$, we propose to transfer the knowledge of the teacher $\mathcal{T}$ to $\mathcal{S}_1$ and $\mathcal{S}_2$ via both the feature information via a novel feature KD bridge and predicted labels via pixel-wise losses. Specifically, for the distillation of predicted label information, we model the loss using the commonly used pixel-wise loss, \eg, $l_1$, which is formulated as follows.  

% \begin{equation}
% \begin{split}
%       \mathcal{L}_{KD}^{pw} = \alpha ||T(x_{aps}) - S_{aps}(x_{aps})||_1 + \\
%       \beta || T(x_{aps}) - S_{ev}(e)||_1
% \end{split}
% \end{equation}

\vspace{-1pt}
\subsection{Affinity Graph KD and Other KD Losses}
\begin{figure}[t!]
    \centering
     \captionsetup{font=small}
    \includegraphics[width=0.98\columnwidth]{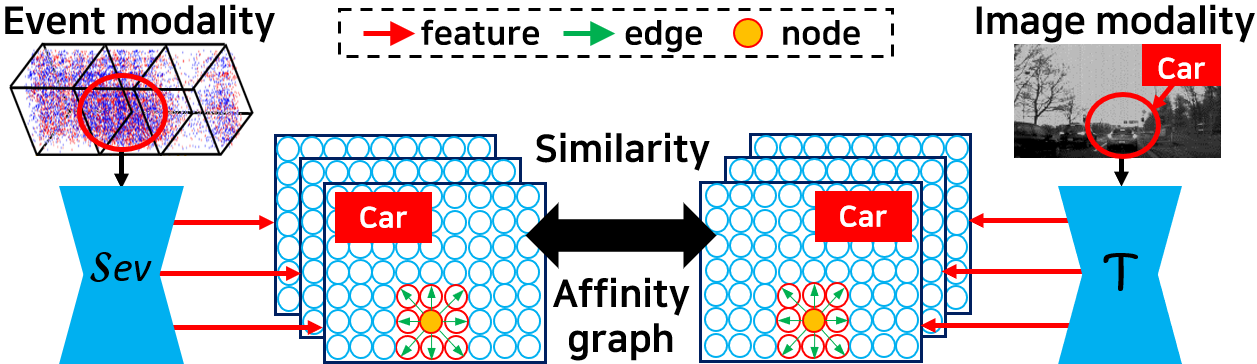}
    \vspace{-8pt}
    \caption{Illustration of the affinity graph distillation loss between the event and image modalities.}
    \label{fig:affinity_graph}
    \vspace{-10pt}
\end{figure}

\vspace{-2pt}
\label{loss_terms}
\noindent \textbf{Affinity Graph (AG) KD.} 
% As ground truth labels are unavailable for event data, using commonly pixel-level loss is insufficient to some extent.
Teacher's features contain constructive knowledge; 
% we It might be possible to apply existing feature-based losses, \eg, \cite{zhao2020knowledge, park2019feed}; 
however, 
% as event and image data belong to two different modalities
due to modality difference, directly matching feature information \cite{zhao2020knowledge, park2019feed, liu2019structured} is impractical. We notice that 
% although they are heterogeneous in modality, 
two modalities share similar labeling contiguity among spatial locations for, \eg, urban scenes. 
% We thus consider distilling the spatial similarity information from the teacher  $\mathcal{T}$ to the students $\mathcal{S}_{aps}$ and $\mathcal{S}_{ev}$.
% To this end, 
Inspired by \cite{liu2019structured,deng2019towards,deng2019towards}, we thus utilize affinity graphs to transfer the instance-level similarity along the spatial locations between two modalities, as shown in Fig.~\ref{fig:affinity_graph}.
% An affinity graph denotes the spatial pair-wise relations between two modalities. 
The node represents a spatial location of an instance (\eg, car), and the edges connected between two nodes represent the similarity of pixels. For events, if we denote the connection range (neighborhood size) as $\sigma$, then nearby events within $\sigma$ (9 nodes in Fig.~\ref{fig:affinity_graph}) are considered for computing affinity contiguity. It is possible to adjust each node's granularity to control the size of the affinity graph; however, as events are sparse, we do not consider this factor. In such a way, we can aggregate top-$\sigma$ nodes according to the spatial distances and represent the affinity feature of a certain node. For a feature map $F \sim \mathbb{R}^{C\times H \times W}$ ($H \times W$ is the spatial resolution and $C$ is the number of channels), the affinity graph contains nodes with $H \times W \times \sigma$ connections. In the two modalities, we denote $A_{uv}^{\mathcal{T}}$ and $A_{uv}^{\mathcal{S}}$ are the affinities between the $u$-th node and the $v$-th node obtained from the teacher and student, respectively, which is formulated as:
% which can be formulated as:
% follows:
\vspace{-3pt}
\begin{equation}
\label{affinity_graph}
    \mathcal{L}_{AG} = \frac{1}{H\times W\times\sigma}\sum_{u\sim R}\sum_{v\sim \sigma}||A_{uv}^{\mathcal{T}} -A_{uv}^{\mathcal{S}}||_2^2 \\[-3pt]
\end{equation}
where $R=\{1,2,\cdots, H \times W \}$ indicates all the nodes in the graph. The similarity between two nodes is calculated from the aggregated features $F_u$ and $F_v$ as $A_{uv}= \frac{F_u^{\intercal} F_v}{||F_u{\intercal}||_2 || F_v||_2}$, 
% \begin{equation}
%   A_{ij}= \frac{F_i^{\intercal} F_j}{||F_i^{\intercal}||_2 || F_j||_2} 
% \end{equation}
where $F_u^{\intercal}$ is the transposed feature vector of $F_u$~\cite{deng2019towards,liu2019structured}. 

\vspace{2pt}
\noindent \textbf{Mutual Distillation (MD).} When APS frames exist, 
% $\mathcal{S}_{ev}$ and $\mathcal{S}_{aps}$ take the event streams and APS frames as inputs.
$\mathcal{S}_{aps}$ indeed can facilitate the learning of $\mathcal{S}_{ev}$. Since $\mathcal{S}_{aps}$ and $\mathcal{S}_{ev}$ start from different initial conditions, they learn different representations, and consequently, their prediction of probabilities can be an effective regularization to each other. We thus let $\mathcal{S}_{aps}$ and $\mathcal{S}_{ev}$ learn from each other's predictions via the KL divergence losses with a temperature parameter $\tau$ \cite{zhang2018deep}.  This helps $\mathcal{S}_{ev}$ converge to better minima for better generalization to test data. The MD loss is formulated as:
\vspace{-3pt}
\begin{equation}
\begin{split}
   \mathcal{L}_{MD} = \mathbb{E}_{e, x_{aps} \sim \mathcal{X}_T}KL \left[\mathcal{S}_{ev}(e) | \mathcal{S}_{aps}(x_{aps}),  \tau \right] +  \\
  \mathbb{E}_{e, x_{aps} \sim \mathcal{X}_T} KL \left[\mathcal{S}_{aps}(x_{aps}) | \mathcal{S}_{ev}(e), \tau \right].  \\[-1pt]
\end{split}
\end{equation}
In summary, the objective $\mathcal{L}$ of EvDistill is as follows : 
\vspace{-3pt}
 \begin{equation}
     \mathcal{L} = \mathcal{L}_{CE} + \mathcal{L}_{BMR} + \lambda_1\mathcal{L}_{DA} + \lambda_2\mathcal{L}_{AG} + \lambda_3\mathcal{L}_{MD} \\[-3pt]
 \end{equation}
 where $\lambda_1$, $\lambda_2$ and $\lambda_3$ are the hyper-parameters.

% \vspace{-3pt} 
\section{Experiment and Evaluation}
% \vspace{-3pt} 

% In this section, we show the experimental results of the proposed framework.
\begin{table*}[t!] 
\renewcommand{\tabcolsep}{16pt}
 \centering
 \footnotesize
 \captionsetup{font=small}
 \caption{Segmentation performance with different event representations and APS images on the test data \cite{alonso2019ev}, measured by MIoU.} 
%  The models are trained on time intervals of $50$ms but tested with $50$ms, $10$ms and $250$ms.}
 \vspace{-9pt}
 \begin{tabular}{l|c|c|c|c|c}
% \toprule[0.8pt]
\hline
 Method & Event Rep.&Use pseudo labels &MIoU [50ms] & MIoU [10ms] & MIoU [250ms]\\
%  \midrule[0.5pt]
\hline
EvSegNet~\cite{alonso2019ev} & 6-channel~\cite{alonso2019ev} & Yes &  54.81 &  45.85  &47.56 \\
Vid2E \cite{gehrig2020video} & EST~\cite{gehrig2019end}& Yes  & 45.48  &30.70 &  40.66\\
% \midrule[0.5pt]
\hline
Ours & Voxel bins (2ch)& No  & 57.16 &  48.68 &  51.23 \\
Ours & Multi-channel &  No & \textbf{58.02} & \textbf{49.21} &\textbf{52.01} \\
% \midrule[0.5pt]
% \hline
\hline 
EvSegNet (APS) & -& Yes & 64.98 &  64.98 &  64.98\\
\hline Ours (APS) & - & No &\textbf{72.63} & \textbf{72.63} & \textbf{72.63}\\
% \hline
\hline
 \end{tabular}
\label{tab:comp_table1}
\vspace{-12pt}
\end{table*}

\subsection{Event-based semantic segmentation}
\vspace{-3pt}
\label{ev_seg_sec}
% In this section, we first evaluate EvDistill  on semantic segmentation task.
% where a event segmentation network is learned by distilling the knowledge from the teacher network on labeled image modality data.
Semantic segmentation is a task that aims to assign a semantic label, \eg, road, car, in a given scene to each pixel. 
% It has been broadly applied to many fields, including pedestrian detection, lane detection for autonomous driving. 
% It has been shown that event camera alone is enough to perform some high-level tasks, such as tracking and SLAM \cite{maqueda2018event, mueggler2017event}; however, it has been less studied regarding semantic segmentation. One reason that hinders the progress on this direction is the lack of qualitative annotated event data \cite{alonso2019ev, gehrig2020video}.
Unlike image data, annotating events requires special processing for the raw event streams. Besides, 
% even when we could have a good way to represent events, 
it is challenging to correctly label the pixels due to the sparsity of events and lacking information (\eg, material and textures).

\vspace{2pt}
\noindent \textbf{Datasets.} We use the publicly available driving scene dataset DDD17 \cite{binas2017ddd17}, which includes both events and APS frames recorded by a DAVIS346 event camera. In \cite{alonso2019ev}, 19,840 APS frames are utilized to generate pseudo annotations(6 classes) based on a pretrained network for events (15,950 for training and 3,890 for test). However, such a way leads to less precise segmentation labels because there is a considerable domain gap between the source data
% APS frames and Cityscapes images, 
and APS images of low resolution and quality. Note that our method does not rely on any annotations of events in training, and the pseudo annotations by \cite{alonso2019ev} for test are only used for evaluation and comparison. 
% we use the pseudo annotations provided by \cite{alonso2019ev}.
% Although \cite{alonso2019ev} utilized the APS frames to generate pseudo labels from a pretrained network on Cityscapes dataset,
% In such a way, the pseudo annotations of DDD17 are synchronized with the APS frames. 
As the events in the DDD17 dataset are very sparse and noisy, we show more results on the driving sequences in the MVSEC dataset \cite{zhu2019unsupervised}, 
collected for the stereo purpose. Moreover, we show qualitative results on the E2Vid driving scene dataset \cite{rebecq2019high} captured by using a Samsung event camera (with higher resolution). 
% and a Huawei phone camera as reference. 
% As there are no semantic annotations for this sequence, we only show the qualitative results for these two datasets. 

\vspace{2pt}
\noindent \textbf{Implementation details.} For each label of DDD17 dataset provided by \cite{alonso2019ev}, we use the events occurred in a 50ms time window before a label for prediction, as done in \cite{alonso2019ev, gehrig2020video}. We also consider the event representation method in \cite{alonso2019ev} for semantic segmentation on the DDD17 dataset, in addition to that described in Sec.~\ref{event_rep}. 
% We consider two event representation methods: two channel event image by splitting both positive and negative events, 2-channel image by merging positive/negative events with corresponding mean and standard deviation and 6-channel representation in  \cite{alonso2019ev}. 
For the event representation on MVSEC and E2VID datasets, we use the method in Sec.~\ref{event_rep}. 
% In \cite{alonso2019ev}, a 6-channel representation of events consists of 3-channel for both positive and negative events. The first channel is simply the histogram of events, namely the number of events accumulated within a certain time interval. The second channel is the mean timestamp (positive and negative) of events and the third channel is the standard deviation (positive and negative) of timestamps (positive and negative).
For the teacher $\mathcal{T}$ and students $\mathcal{S}_{ev}$ and $\mathcal{S}_{aps}$, we adopt a segmentation network \cite{chen2018encoder}. 
% \noindent \textbf{Evaluation metrics}
We use the following metric to evaluate the performance.  The \textit{intersection of union} (IoU) score is calculated as the ratio of intersection and union between the GT mask and the predicted segmentation mask for each class. We use the \textit{mean IoU} (MIoU) to measure the effectiveness, as done in \cite{chen2017deeplab,chen2018encoder}. The segmentation maps are with 6 classes, as done in~\cite{alonso2019ev}.
% We also report the \textit{pixel accuracy}, which is the ratio of the pixels with the correct semantic labels to the overall pixels. 
For more implementation details (\eg, training), refer to the suppl. material.

\vspace{-6pt}
\subsubsection{Evaluation on DDD17 dataset}
\begin{figure}[t!]
        \centering
         \captionsetup{font=small}
        \includegraphics[width=\columnwidth]{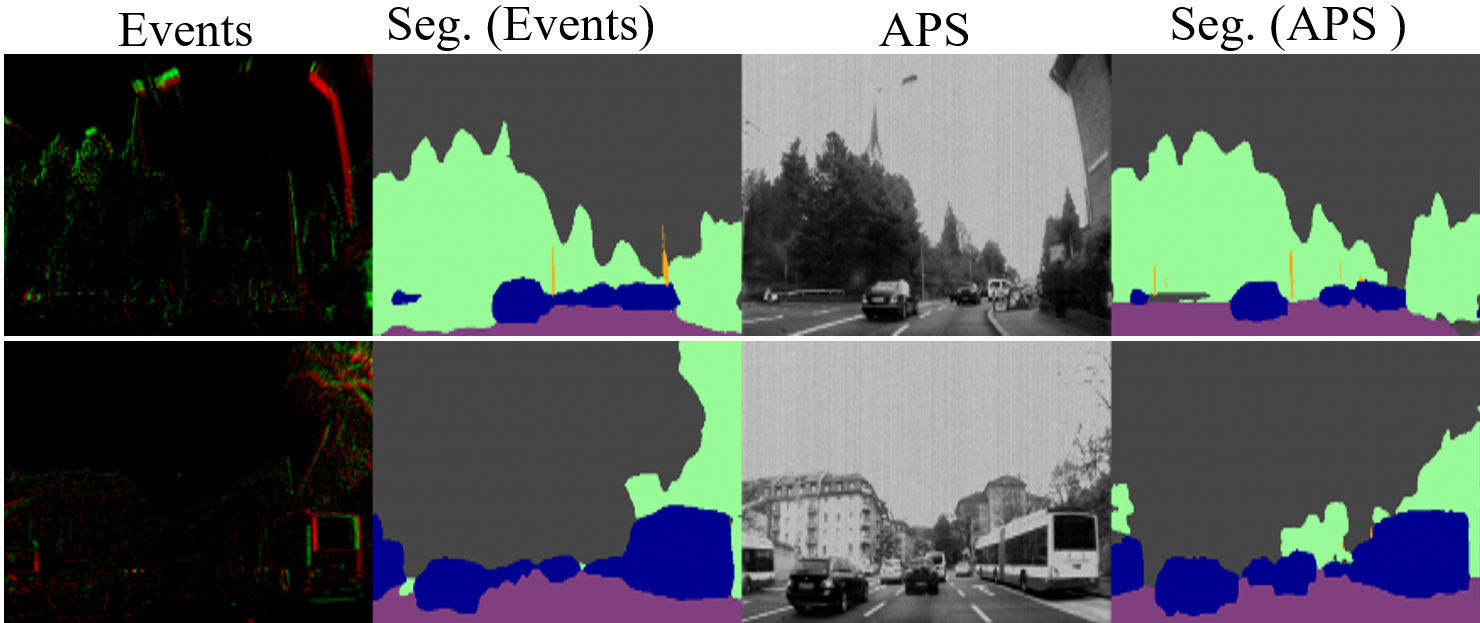}
        \vspace{-18pt}
        \caption{Semantic segmentation results of urban driving scenes on DDD17 test dataset (gray: background; green: vegetation; blue: vehicle; violet: street; yellow: object).
        % From left to right, events, segmentation results on events, APS frames, segmentation results on APS frames.
        }
        \label{fig:event_seg}
        \vspace{-10pt}
    \end{figure}

\vspace{-3pt}
\noindent \textbf{Comparison.} We first present the experimental results on the DDD17 dataset~\cite{alonso2019ev}. We evaluate our method on the test set and vary the window size of events between 10, 50, and 250ms, as done in ~\cite{alonso2019ev}. The quantitative and qualitative results are shown in Table~\ref{tab:comp_table1} and Fig.~\ref{fig:event_seg}.  We compare our method with two existing methods, EvSegNet~\cite{alonso2019ev} and Vid2E~\cite{gehrig2020video} that uses synthetic version of DDD17 data. It turns out that, without using labels, EvDistill significantly improves the segmentation results on events and surpasses the existing methods with around $7\%$ increase in MIoU using a multi-channel event representation. Segmentation using the voxel grid representation is slightly less effective than that of using the multi-channel representation. Meanwhile, on the time interval of 10ms and 250ms, EvDistill also shows a significant increase of MIoU by around $7.5\%$ and $9.5\%$ than those of EvSegNet, respectively.  

We also demonstrate that our method significantly enhances the semantic segmentation performance on the APS frames in Table~\ref{tab:comp_table1}. Quantitatively, without resorting to the pseudo labels, our method surpasses EvSegNet by a large margin with around $12\%$ increase of MIoU. This reflects that our method dramatically minimizes the domain gap between APS frames and labeled source data and distills the knowledge from the teacher network to learn a student network $\mathcal{S}_{aps}$ showing better segmentation capability. The results on both events and APS frames indicate that our method successfully distills the knowledge from the teacher network learned on the labeled image modality data for tackling the challenges of unpaired and unlabeled events.

\vspace{2pt} 
\noindent \textbf{High dynamic range (HDR).} HDR is one distinct advantage of event cameras. Even when APS frames are ill-exposed, events capture the intensity changes. We show the student network $\mathcal{S}_{ev}$ shows promising performance in the extreme condition.  Fig.~\textcolor{red}{4} of the suppl. material shows the qualitative results. The APS frames are over-exposed, thus the student network $\mathcal{S}_{aps}$ fails to segment the urban scenes; however, the events capture the scene details, and the student network $\mathcal{S}_{ev}$ shows convincing segmentation results.  

% \begin{table}[t!] 
% \renewcommand{\tabcolsep}{5.0pt}
%  \centering
%  \footnotesize
%  \captionsetup{font=small}
%  \caption{A comparison of the unsupervised domain generalization with EvSegNet}
%  \begin{tabular}{l|c|c|c}
% \toprule[0.8pt]
%  Method & Use GT labels & Acc. [50ms] & MIoU [50ms] \\
%  \midrule[0.5pt]
% EvSegNet (APS)\cite{alonso2019ev} &  Yes & 89.76 & 64.98\\
% \hline
% % Ours (APS) &  No & & 68.12\\
% Ours (APS) &  No &96.56 & 69.74\\
% \bottomrule[0.8pt]
% % \bottomrule[1pt]
%  \end{tabular}
% \label{tab:comp_table_aps}
% \vspace{-10pt}
% \end{table}

% \noindent \textbf{Semantics-oriented Image reconstruction}
% \noindent \textbf{High dynamic range.} One of the significant advantage of event camera is its high dynamic range. Even when frame images are over/under exposed, event streams can capture the intensity changes. Here, we show that our method shows a great potential for event-based semantic segmentation when frame images fail to work. The qualitative results are shown in Fig.~\ref{fig:event_seg}.

% \noindent \textbf{Discussion:} 
\vspace{-10pt}
\subsubsection{Evaluation on MVSEC dataset}
\begin{figure}[t!]
        \centering
         \captionsetup{font=small}
        \includegraphics[width=\columnwidth]{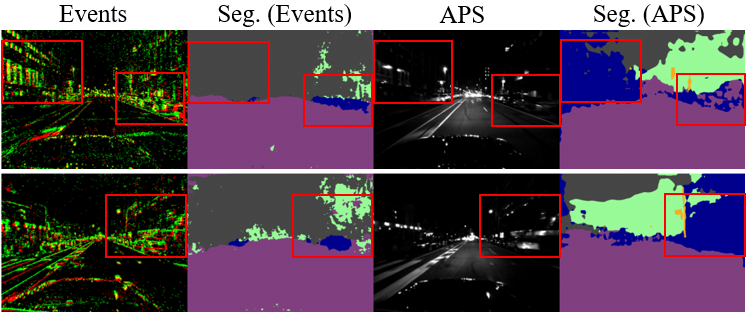}
        \vspace{-20pt}
        \caption{Semantic segmentation results of low-light scenes on MVSEC dataset (gray: background; green: vegetation; blue: vehicle; violet: street; yellow: object).  
        % From left to right, events,  seg. results on events, APS frames, seg. results on APS frames.
        }
        \label{fig:event_seg_stereo}
        \vspace{-12pt}
    \end{figure}

\vspace{-5pt}
% Indeed, the raw events in  DDD17 dataset are very sparse and noisy. Even with the 6-channel representation in EvSegNet, there lack events for the critical objects (\eg, people, car) in the driving scenes. Meanwhile, the APS images are with noise and unpleasant artifacts. Both the low quality of  events and APS frames may degrade the performance of semantic segmentation. 
We further present the experimental results on the MVSEC dataset \cite{zhu2019unsupervised}, which contains various driving scenes for stereo estimation. 
% The dataset contains various driving scenes and the events are of relatively higher quality than DDD17 dataet. 
We use the `outdoor\_day2' sequence and divide the data into training and test sets. We remove the redundant sequences, such as vehicles stopping in the traffic lights, etc. We also use the night driving sequences to show the advantage for HDR. For the details of dataset preparation and more visual results, refer to the suppl. material. To quantitatively evaluate our method, we also utilize the APS frames to generate pseudo labels, similar to \cite{alonso2019ev}, as our comparison baseline. The qualitative and quantitative results are shown in Fig.~\ref{fig:event_seg_stereo} (also see Fig.~\textcolor{red}{2} of the suppl. material) and Table~\ref{tab:comp_tab_stereo}. In Fig.~\ref{fig:event_seg_stereo}, we mainly show the results in the low-light condition. It is evident that, although the student network $\mathcal{S}_{aps}$ fails to work on APS frames, where most pixels in the red boxes are wrongly classified in the 4th column (\eg, building and trees misclassified as vehicles), events capture scene information better and provide better segmentation performance in low-light condition. Compared with the baseline in Table~\ref{tab:comp_tab_stereo}, our method significantly surpasses it by a noticeable margin with a $8.8\%$ increase of MIoU on the events and a $11.2\%$ increase on the APS frames.

% stack the events using the representation method depicted in Sec.~\ref{event_rep}. 
\begin{table}[t!] 
\renewcommand{\tabcolsep}{17.0pt}
 \centering
 \footnotesize
 \captionsetup{font=small}
 \caption{Segmentation performance of events and APS images on the MVSEC dataset \cite{zhu2018multivehicle}, measured by MIoU. 
 The baseline is trained by using the pseudo labels made from the APS images. }
\vspace{-8pt}
 \begin{tabular}{l|c|c}
% \toprule[0.7pt]
\hline
 Method & Use pseudo labels & MIoU\\
% \midrule[0.5pt]
\hline
Baseline (Events) & Yes &  50.53 \\
Ours(Events) & No & \textbf{55.09}\\
\hline
Baseline (APS) & Yes &  61.93\\
Ours (APS) & No & \textbf{68.85}\\
\hline
 \end{tabular}
\label{tab:comp_tab_stereo}
\vspace{-5pt}
\end{table}

\vspace{-10pt}
\subsubsection{Evaluation on E2VID dataset}
\begin{figure}[t!]
        \centering
         \captionsetup{font=small}
        \includegraphics[width=\columnwidth]{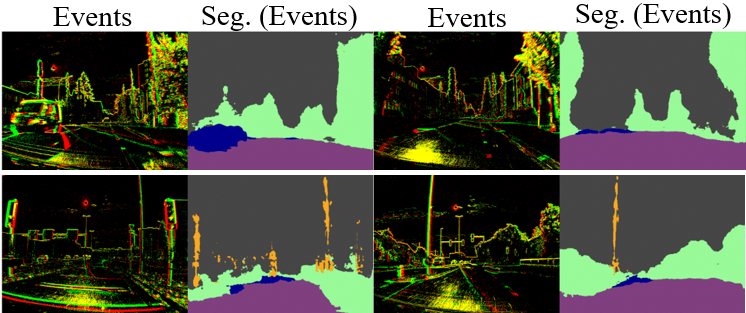}
        \vspace{-20pt}
        \caption{Qualitative results on E2Vid dataset (gray: background; green: vegetation; blue: vehicle; violet: street; yellow: object).  
        % The 1st and 3rd columns are the events, 2nd and 4th columns are the seg. results.
        }
        \label{fig:event_seg_e2vid}
        \vspace{-8pt}
    \end{figure}

\vspace{-5pt}
We also present the experimental results on the E2Vid driving dataset \cite{rebecq2019high}. We followed the DDD17 in \cite{alonso2019ev} to split E2VID dataset. We mainly use `sun2' and `sun4' sequences where we select around 4K event images as the training set, and the remained 400 as the test set. We also test on 400 event images from the `street' sequence. 
% For more details of the dataset, refer to the suppl. material. 
As there are no GT annotations for events, we only show the qualitative results, as shown in Fig.~\ref{fig:event_seg_e2vid} and Fig.~\textcolor{red}{3} of the \textit{suppl. material}. The student network $\mathcal{S}_{ev}$ can segment the moving objects, \eg, vehicles, pedestrians. Meanwhile, it also successfully segments the complex objects with no motion blur,\eg, tree branches, traffic lights. Compared with the events (346x260) in DDD17 and MVSEC datasets, the events in E2Vid are in a higher resolution (640x480). In Fig.~\ref{fig:event_seg_e2vid}, it seems that the network trained on these events better segments small objects, \eg, vehicles in the remote location, traffic lights, etc.   
% higher than that of the events in a lower resolution. 
Although events contain less visual information than the image data, it is advantageous for segmenting the fast moving objects and HDR scenes. 

% We have demonstrated that, by distilling knowledge from the teacher network trained with the annotated RGB images, the event network $\mathcal{S}_{ev}$ can successfully segment the input event streams. Here, we demonstrate that our KD method also generalize to the condition where traditional intensity (APS) images suffer from the motion blur or over-/under-exposure. By learning the knowledge from the teacher, the EventNet is  well trained to abundant event inputs. 

\vspace{-3pt}
\subsection{Event-based object recognition}
\label{cls_kd}
\begin{figure}[t!]
    \centering
     \captionsetup{font=small}
    \includegraphics[width=.98\columnwidth]{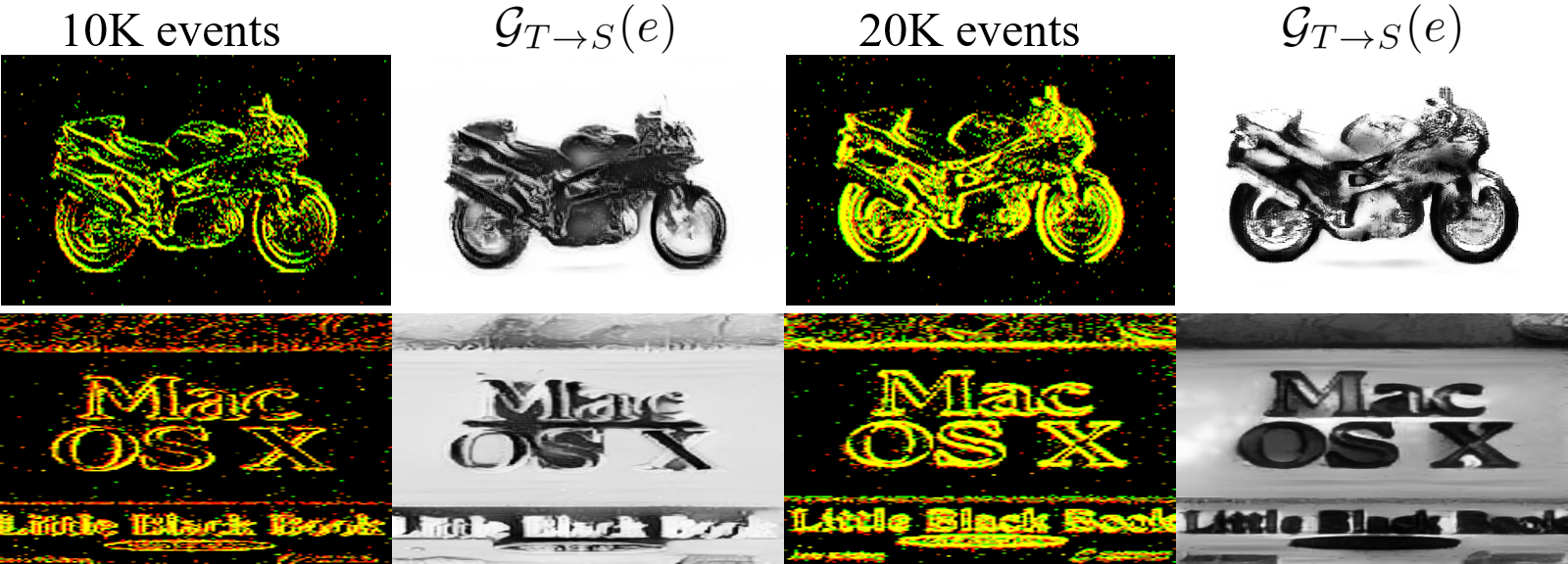}
    \vspace{-8pt}
    \caption{Generated images with $\mathcal{G}_{T \to S}$ (2nd and 4th columns based on 10K and 20K events) from the target events.}
    \label{fig:ev2im}
    \vspace{-8pt}
\end{figure}

\begin{figure}[t!]
    \centering
     \captionsetup{font=small}
    \includegraphics[width=.98\columnwidth]{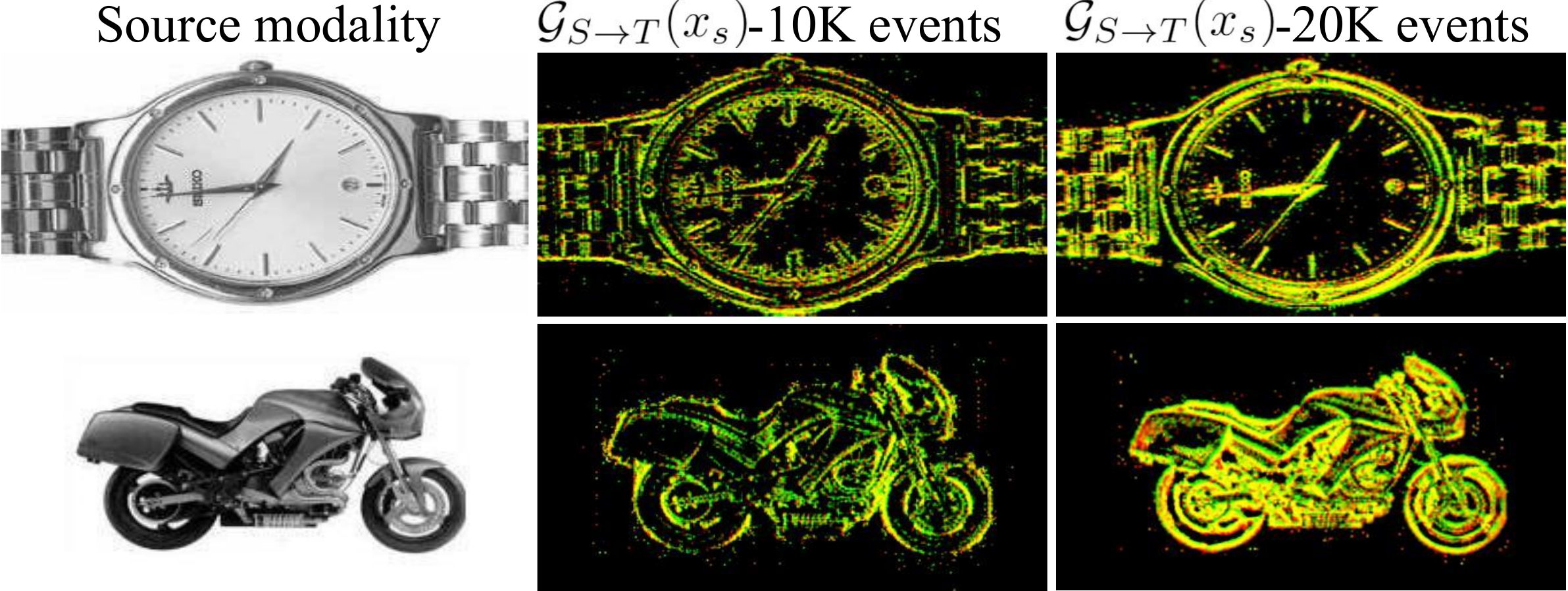}
    \vspace{-8pt}
    \caption{Generated events with $\mathcal{G}_{S \to T}$ (10K and 20K events in the 2nd and 3rd columns) from the source images.}
    \label{fig:im2events}
    \vspace{-12pt}
\end{figure}

\vspace{-3pt}
% Object recognition or classification from event streams is a pattern recognition task that has been actively studied in recent years. Event-based object recognition overcomes the challenges that standard cameras have due to their high dynamic range, no motion and low latency. 
We further demonstrate that our method can be also flexibly applied to object recognition. We use the benchmark N-Caltech101 dataset \cite{orchard2015converting}. This dataset is an event-based version of the well-known Caltech101 \cite{fei2006one}. Note that the event data in the N-Caltech dataset and the original images in the Caltech dataset are not matched. As the dataset only provides events captured by an event camera, without APS frames, the student network $\mathcal{S}_{aps}$ depicted in Fig.~\ref{fig:kd_method} is removed in this case. 
% This makes it impossible to achieve semantics-enhanced image reconstruction from events in a supervised manner. 
% We instead study a slightly different scenario when APS images are not given, as shown in Fig.~\ref{fig:cls_kd}. 
To bridge both modalities, we explore the source images (\eg, the images from Caltech dataset and other images) and learn the bidirectional modality reconstruction (BMR) module in an unsupervised manner.
% (see details in Sec.~\textcolor{red}{4} of suppl. material). 
% as described in Sec.~\ref{data_augmentation}.
% , as inspired by recent domain transfer methods \cite{wang2020eventsr, zhu2019unsupervised,liu2017unsupervised}. 
Note that we assume that the labels of event data are unknown and the labels of source images are given. We utilize a teacher classifier $\mathcal{T}$ pretrained on the source
images and distill the knowledge to the student classifier $\mathcal{S}_{ev}$. Interestingly, EvDistill translates the source images (with labels) to events also with the same labels (Fig.~\ref{fig:im2events}). We then utilize the generated events and target events to train the student classifier $\mathcal{S}_{ev}$. Meanwhile, the generated images (Fig.~\ref{fig:ev2im}) and source images are used to get recognition information from the teacher $\mathcal{T}$, such that the student can learn the distilled knowledge from the teacher via the KD losses. From the experiments, we show there is a significant performance boost (see Table~\ref{tab:comp_tab_cls}) with our method. 

% In our evaluation, we mainly use the benchmark N-Caltech101 (Neuromorphic-Caltech101) dataset \cite{orchard2015converting}, which is the event-based version of the well-known Caltech101 \cite{fei2006one}. The Caltech101 dataset is one of the imbalanced dataset that brings a major problem for training the event network. The N-caltech101 dataset consists of 8709 event sequences from 101 object classes, and each sequence holds a duration of 300ms. The N-caltech dataset is captured by mounting an event camera and having the camera saccadicly move when it views the example images from Caltech dataset shown in the screen of a projector. 
% The proposed KD framework is oriented for the condition where both event streams and APS frames from event cameras are given. 
% In particular, we provide an analysis of a slightly different scenario when APS images are not given. 
% In Fig.~\ref{fig:kd_method}, The proposed KD framework is oriented for the condition where both event streams and APS frames from event cameras are given. However, the framework can also be flexibly adapted when APS frames are not provided by some event cameras (\eg, DVS ). 

% \begin{figure}[t!]
%     \centering
%     \includegraphics[width=\columnwidth]{figures/cls.png}
%       \vspace{-20pt}
%     \caption{Unsupervised semantics-enhanced image reconstruction of cross-model KD for event-based classification.}
%     \label{fig:cls_kd}
  
% \end{figure}

\begin{table}[t!] 
\renewcommand{\tabcolsep}{4.8pt}
 \centering
 \footnotesize
 \captionsetup{font=small}
 \caption{A comparison of object recognition performance with existing methods on N-Caltech dataset.}
 \vspace{-8pt}
 \begin{tabular}{l|c|c|c}
\toprule[0.7pt]
 Method & Training data & Use GT labels & Test score \\
 \midrule[0.5pt]
HATS~\cite{sironi2018hats} & Real events & Yes & 0.642 \\
HATS-ResNet34 & Real events &  Yes & 0.691\\
RG-CNN &Real events & Yes & 0.657 \\
EST~\cite{gehrig2019end}&Real events & Yes & 0.817 \\
E2VID~\cite{rebecq2019high} & Intensity images & Yes & 0.866\\
VID2E~\cite{gehrig2020video} & Synthetic events & Yes &  0.807 \\
\midrule[0.5pt]
Ours-20K events & Real events & No & 0.896 \\
Ours (fine-tune) & Real events & No & \textbf{0.902} \\
% Ours-2bins & Real events & Yes & \textbf{0.915}\\
\bottomrule[0.7pt]
% \bottomrule[1pt]
 \end{tabular}
\label{tab:comp_tab_cls}
\vspace{-8pt}
% \vspace{-20pt}
\end{table}

% \begin{figure*}[t!]
% % %\vspace{-10pt}
% \begin{center}
% \renewcommand{\tabcolsep}{1pt}
% \begin{tabular}{@{}c@{}}

% %  \includegraphics[width=\textwidth]{figures/gen_a2b_test_00930000.jpg} \\
% %   Event to images \\
%   \includegraphics[width=\textwidth]{figures/gen_b2a_train_current.jpg}\\
%   Image to event \\
%  \end{tabular}
% % \captionsetup{font=small}
% \caption{Visual examples of event to image and image to events synthesis on N-calteach dataset.}
% \label{fig:id_attack}
% \end{center}
% \end{figure*}

\vspace{1pt}
\noindent \textbf{Implementation Details.} 
% As only events are provided in the N-Caltech dataset and original color images in Caltech dataset are not matched with events, we design the SEIS module in an unsupervised manner. The adapted SEIS module enable to generate intensity images from events. Meanwhile, it also enables generating events from real-world events. 
For the teacher and student classifiers, we use the ResNet34, as used in other works \cite{gehrig2020video, rebecq2019high, gehrig2019end}. We use the loss functions defined in Eq.~(\ref{kl_div}) and Eq.~(\ref{affinity_graph}), and also cross-entropy loss. 
We use Adam optimizer with the learning rate of 1e-4. For event representation, we use the stacking method described in Sec.~\ref{event_rep}.  Due to the lack of space, more details of the implementation are provided in the supplementary material.

\vspace{2pt}
\noindent \textbf{Experimental results.} The qualitative results for image generation are shown in Fig.~\ref{fig:ev2im}. The first and third columns show the stacked 10K and 20K events, accompanied by the generated images in the 2nd and 4th columns. As can be visually seen, even without supervision, realistic images are generated from the target modality. When more events are accumulated, better-reconstructed images are obtained. Correspondingly, Fig.~\ref{fig:im2events} shows the generated events from the source modality. The 1st column shows the source images, and the 2nd and 3rd columns are the generated 10K and 20K events, respectively. It is clearly shown that the quality of the generated events is improved based on the target events. Regarding the results of object recognition, we compare with the SoTA optimization-based methods, HATS \cite{sironi2018hats} and its ResNet34-based result. Meanwhile, we compare with the recent DL-based methods, EST \cite{gehrig2019end}, E2Vid \cite{rebecq2019high} (using generated images) and Vid2E \cite{gehrig2020video} (using synthetic events).  Table~\ref{tab:comp_tab_cls} shows the quantitative
results. Even when labels are not used for event data, our method surpasses the existing methods with a significant margin. For instance, compared with HATS-Resnet34, our method achieves more than $20\%$ higher accuracy. When compared with E2Vid, our method also achieves better results (0.896 vs 0.866). When tuning parameters via self-ensemble in training, EvDistill further improves the performance and achieves higher accuracy (0.902 vs 0.896).

\vspace{-5pt}
\section{Ablation Study and Analysis}
\begin{figure}[t!]
    \centering
    \captionsetup{font=small}
    \includegraphics[width=\columnwidth]{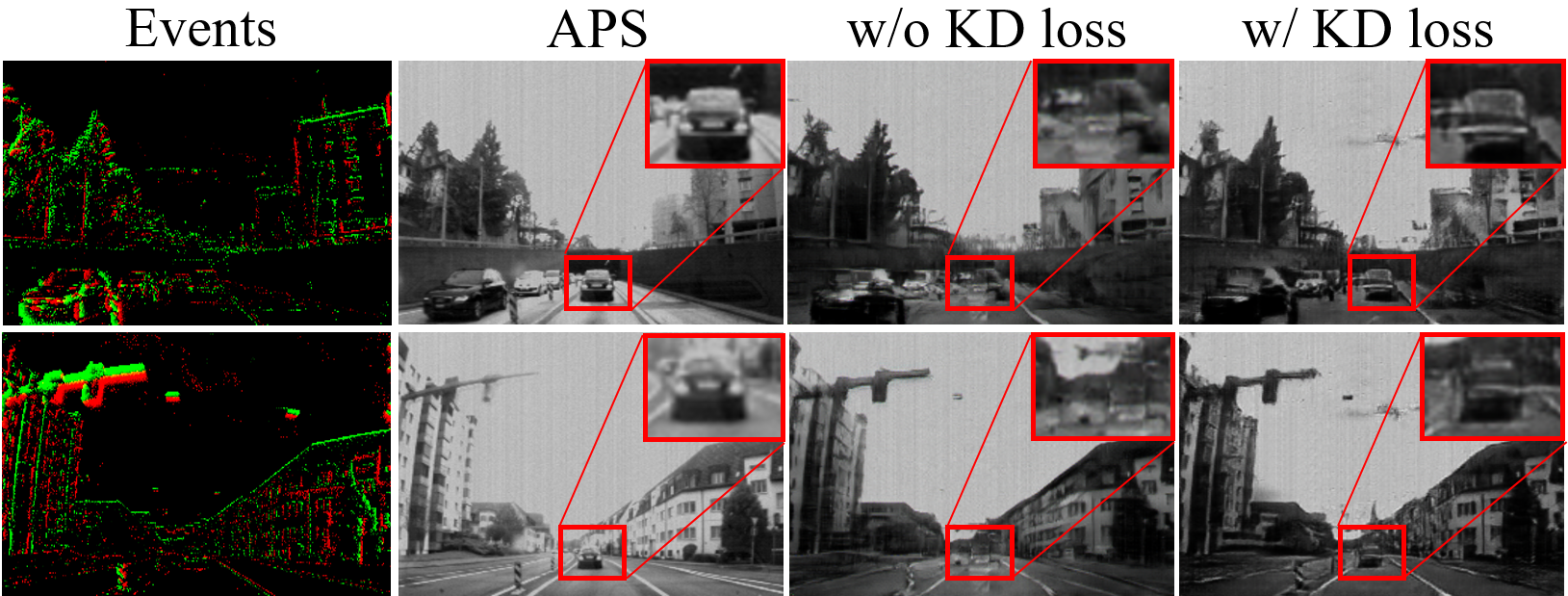}
    \vspace{-20pt}
    \caption{Visual results of the end-to-end target-to-source reconstruction with (w/) and without (w/o) KD loss.}
    \label{fig:semantic_image}
    \vspace{-14pt}
\end{figure}

\vspace{-4pt}
\noindent \textbf{Modality reconstruction.} 
We show that EvDistill also enhances the target-to-source reconstruction by leveraging the proposed dynamic semantic consistency (DSC) KD loss.
% We also analyze and discuss the experimental results for semantics-enhanced image  reconstruction, 
The qualitative and quantitative results are shown in Fig.~\ref{fig:semantic_image} and Table~\ref{tab:comp_im_rec}. In contrast to the reconstructed images without KD loss (2nd column), our method fully exploits the semantic information and successfully restores the textural and material details, \eg, cars in the cropped patches (3rd column) in Fig.~\ref{fig:semantic_image}. The effectiveness can be further verified from Table~\ref{tab:comp_im_rec}, where the images generated with the KD loss show higher performance for semantic segmentation.
% As the events are relatively sparse, it is difficult to restore the objects, which are important for segmentation, with only pixel-wise loss. Although the adversarial loss helps generate more realistic textural details; however, it can not enhance the semantic information. 
The proposed EvDistill helps recover the semantic information in the generated images and improves semantic segmentation quality on these images. 
\begin{table}[t!] 
\renewcommand{\tabcolsep}{25.0pt}
 \centering
 \footnotesize
 \captionsetup{font=small}
 \caption{Comparison of segmentation results on the target to source reconstruction with and without knowledge distillation.}
\vspace{-9pt}
 \begin{tabular}{l|c}
% \toprule[0.7pt]
\hline
 Method & Mean IoU \\
\midrule[0.5pt]
w/o knowledge distillation  & 43.64  \\
\hline
w/ knowledge distillation & \textbf{45.12}\\
\hline
% \bottomrule[0.7pt]
% \bottomrule[1pt]
 \end{tabular}
\label{tab:comp_im_rec}
\vspace{-8pt}
\end{table}

\vspace{2pt}
\noindent \textbf{The effectiveness of affinity graph KD.} To further validate the  effectiveness of the proposed affinity graph (AG) KD for cross-modal learning, we compare with the general feature-level KD losses, \eg, FitNets \cite{romero2014fitnets}, AT \cite{zagoruyko2016paying}, and FT \cite{kim2018paraphrasing}. The results are shown in Table~\ref{tab:comp_tab_ag}. As FitNets, AT and FT are all targeted to directly minimizing feature difference between the teacher and student under the same modality (\eg, image) data, they show relatively poor performance on the cross-modal learning problem. Instead of directly matching features, the proposed AG loss better tackles the spatial contiguity of instances between the two modalities and shows better performance on the end-tasks.    

\vspace{2pt}
\noindent \textbf{The effectiveness of distillation.}
% We also conduct ablation studies on the KD loss functions. 
We look into the effect of enabling and disabling different components of the  proposed EvDistill. The experiments were conducted on the semantic segmentation task with the DDD17 dataset. In Table~\ref{tab:comp_tab_ablation}, the results of different settings for the student network are listed. Our baseline framework consists of a teacher $\mathcal{T}$ and two student networks.  From Table~\ref{tab:comp_tab_ablation}, we can see that KD can improve the performance of both student networks. Moreover, each KD scheme leads to higher test scores. This implies that the KD schemes make a complementary contribution to learning better student network. Furthermore, it is shown that
% we can see that 
the distribution matching KD scheme better matches the structural similarities between the modalities, leading to higher scores and better quality.  On the other hand, the proposed BMR approach (with (w/) BMR) also contributes to enhancing the learning of the event-based segmentation network. When the student network $\mathcal{S}_{aps}$ is added, it also benefits the learning of $\mathcal{S}_{ev}$ optimized by the mutual learning KD and distribution matching KD losses. 

\vspace{2pt}
\noindent \textbf{KD with only events and APS frames.} Although the paired events and APS frames are without annotated labels, one naive way might be to utilize them in EvDistill without exploring source data and the BMR module. We study this baseline by feeding the events to the student $\mathcal{S}_{ev}$ and the APS frames to the teacher $\mathcal{T}$ for semantic segmentation on the MVSEC dataset. We apply the proposed distribution adaptation scheme and the affinity graph loss to distill knowledge to the student $\mathcal{S}_{ev}$.  The experimental results show that it achieves less plausible MIoU (52.10 vs. 55.09) than the proposed framework as there is a domain gap with the source data used to train the teacher network. The APS frames are of low quality, which degrades the performance. 
% \textit{More analysis of EvDistill is in the suppl. material}.
% Our baseline framework consists of a teacher $\mathcal{T}$, a student network $\mathcal{S}_{ev}$, and reconstruction network $\mathcal{G}$. When APS frames exist, we add one more student $\mathcal{S}_{aps}$. We then study the effectiveness of affinity graph KD loss, mutual KD loss. 

\begin{table}[t!] 
\renewcommand{\tabcolsep}{6.0pt}
 \centering
 \footnotesize
 \captionsetup{font=small}
 \caption{A comparison of affinity graph KD with existing feature KD methods on DDD17 dataset.}
  \vspace{-10pt}
 \begin{tabular}{l|c|c|c|c|c}
% \toprule[0.7pt]
\hline
 Metric & Event Rep.& FitNet \cite{romero2014fitnets} & AT \cite{zagoruyko2016paying} & FT \cite{kim2018paraphrasing} & AG \\
\hline
MIoU & Voxel-2ch & 55.09 & 55.60 & 55.51 & 57.16 \\
% \bottomrule[0.7pt]
\hline
% \bottomrule[1pt]
 \end{tabular}
\label{tab:comp_tab_ag}
\vspace{-5pt}
\end{table}

\begin{table}[t!] 
\renewcommand{\tabcolsep}{6.0pt}
 \centering
 \footnotesize
 \captionsetup{font=small}
 \caption{The effect of different components of EvDistill with a  multi-channel event representation. PI: pixel-wise KD, AG: affinity graph, DM: distribution adaptation, ML: mutual learning.}
 \vspace{-9pt}
 \begin{tabular}{l|c|c}
\toprule[0.7pt]
 Method & Use pseudo labels & Performance (MIoU) \\
% \midrule[0.5pt]
\hline
PI  & No & 55.10 \\
PI + AG  & No & 56.59\\
PI + AG + DM & No & 57.86\\
PI + AG + DM + ML & No & 58.02\\
\hline
w/o BMR  & No & 56.25\\
w/ BMR & No & 57.40 \\
w/ BMR + $\mathcal{S}_{aps}$ & No & 58.02 \\
% \bottomrule[0.7pt]
\hline
% \bottomrule[1pt]
 \end{tabular}
\label{tab:comp_tab_ablation}
\vspace{-12pt}
\end{table}

\vspace{-7pt}
\section{Conclusion, Limitations and Future Work}
\vspace{-5pt}
This paper proposed EvDistill to learn a student on the unpaired and unlabeled events by distilling the knowledge from a teacher trained with labeled images. As no paired modality data with common labels exist, we proposed a BMR module to bridge both modalities. We also proposed a distribution adaptation scheme to match the distributions of two modalities.  Besides, a novel graph affinity KD was proposed to enhance the KD performance. The experiments on two end-tasks demonstrate the effectiveness of our method. Our work has some limitations. First, the type of event data (\eg, urban driving) needs to be close to the labeled source data. Second, as deep network is used, inference latency is inevitable. As EVDistill is a general approach, which tackles the problem caused by limited labels in the target modality. Thus, it can be flexibly applied to any other data, such as thermal and depth camera data in the future work.

\vspace{3pt}
\noindent \textbf{{Acknowledgement}} 
\small{This work was supported by the National Research Foundation of Korea (NRF) grant funded by the Korea government (MSIT) (NRF-2018R1A2B3008640) and Institute of Information \& Communications Technology Planning \& Evaluation(IITP) grant funded by the Korea government(MSIT) (No.2014-3-00123, Development of High Performance Visual BigData Discovery Platform for Large-Scale Realtime Data Analysis).}
% In this paper, we proposed a novel approach to learn event-based prediction tasks. To overcome the difficulty of obtaining labels and the painful domain gap between synthetic events and real-world events, we propose to distill the knowledge of real-world labeled data (RGB modality) to the unlabeled event data (event modality). As a potential, we also bridge the reconstructed intensity images from events with the APS images in an end-to-end  learning manner to further improve the performance. Experimental results show our method not only empowers the prediction task but also improve the quality of image reconstruction.
% In the further, we aim to apply the propose method to other tasks, \eg, object detection as it is worthwhile to show the potential of event cameras for practical applications, \eg, autonomous driving.
% Moreover, we also aim to explore other event representation methods, such as  EST \cite{gehrig2019end}, to improve the performance of learning.
% \subsection{Object detection}

% {\small
% \bibliographystyle{ieee_fullname}
% \bibliography{egbib}
% }

% \clearpage
{\small
\bibliographystyle{ieee_fullname}
\bibliography{egbib}
}

\end{document}